\documentclass[acmtog]{acmart}

\copyrightyear{2024}
\acmYear{2024}
\setcopyright{rightsretained}
\acmConference[SA Conference Papers '24]{SIGGRAPH Asia 2024 Conference Papers}{December 3--6, 2024}{Tokyo, Japan}
\acmBooktitle{SIGGRAPH Asia 2024 Conference Papers (SA Conference Papers '24), December 3--6, 2024, Tokyo, Japan}
\acmDOI{10.1145/3680528.3687583}
\acmISBN{979-8-4007-1131-2/24/12}

\graphicspath{{figures/}{pictures/}{images/}{./}}
\AtBeginDocument{%
}

\usepackage{multirow}

\begin{document}

\title{High-quality Animatable Eyelid Shapes from Lightweight Captures}

\author{Junfeng Lyu}
\affiliation{
	\institution{School of Software and BNRist, Tsinghua University}
	\city{Beijing}
	\country{China}
}
\email{ljf19@mails.tsinghua.edu.cn}

\author{Feng Xu}
\affiliation{
	\institution{School of Software and BNRist, Tsinghua University}
	\city{Beijing}
	\country{China}
}
\email{feng-xu@tsinghua.edu.cn}

\renewcommand{\shortauthors}{Junfeng Lyu and Feng Xu}

\begin{abstract}
	High-quality eyelid reconstruction and animation are challenging for the subtle details and complicated deformations. Previous works usually suffer from the trade-off between the capture costs and the quality of details. 
	In this paper, we propose a novel method that can achieve detailed eyelid reconstruction and animation by only using an RGB video captured by a mobile phone. 
	Our method utilizes both static and dynamic information of eyeballs (e.g., positions and rotations) to assist the eyelid reconstruction, cooperating with an automatic eyeball calibration method to get the required eyeball parameters. 
	Furthermore, we develop a neural eyelid control module to achieve the semantic animation control of eyelids. 
	To the best of our knowledge, we present the first method for high-quality eyelid reconstruction and animation from lightweight captures. 
	Extensive experiments on both synthetic and real data show that our method can provide more detailed and realistic results compared with previous methods based on the same-level capture setups. 
	The code is available at \url{https://github.com/StoryMY/AniEyelid}.
\end{abstract}



\begin{teaserfigure}
	\centering
	\includegraphics[width=1.0\linewidth]{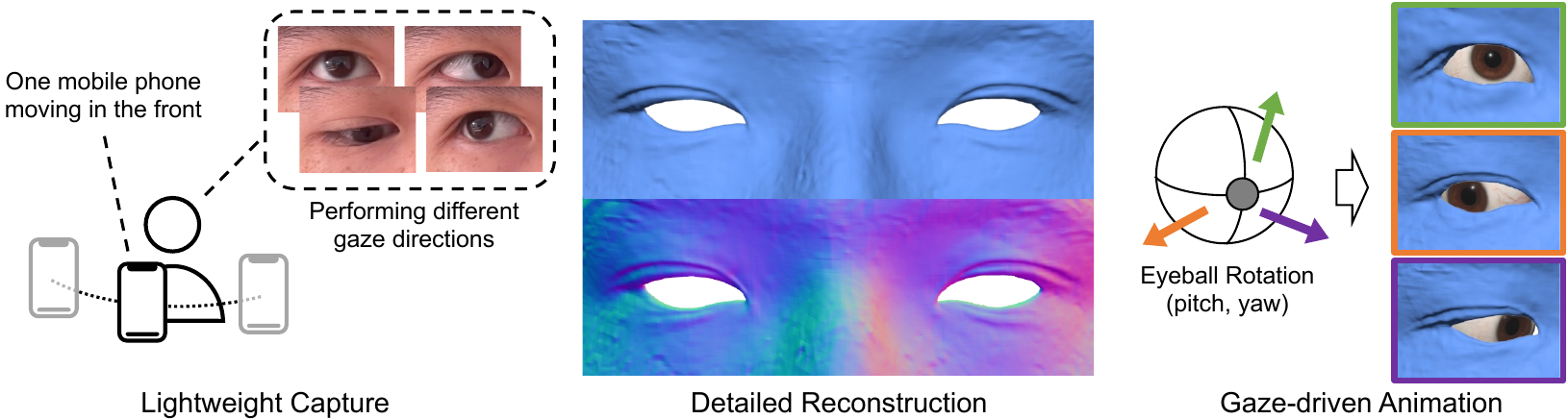}
	\caption{Our approach can leverage lightweight capture to achieve detailed reconstruction and gaze-driven animation of the eyelids.}
	\label{fig:teaser}
\end{teaserfigure}

\maketitle

\section{Introduction}
Realistic digital humans are widely used in computer games, films, and metaverse. As ``the windows to the soul'', eyes play a key role in digital humans, which are significant for creating vivid animation and conveying mental activities. However, although face reconstruction and animation have received much attention from the community, the quality of eyelids still needs to be improved to overcome the uncanny valley.

Obtaining high-quality eyelids is challenging in two aspects. First, the shape of eyelids varies across identities and has subtle details that are hard to capture. Second, the dynamics of eyelids are also very complicated as complex non-rigid deformations like skin folding and unfolding often occur during the movement.

Due to these difficulties, previous methods \cite{wood20163d,wood2016learning,kerbiriou2022detailed} have to make a trade-off between the capture costs and the detail quality. Typically, in order to reconstruct detailed eyelids, \citet{bermano2015detailed} propose a spatio-temporal eyelid reconstruction method that depends on a multi-camera setup. However, this method does not support animating reconstructed eyelids, and the requirement of a high-cost system strictly limits its daily applications. 
On the other hand, based on the pre-designed blendshapes, \citet{wen2017real} reconstructs eyelids by a single RGB-D camera while retaining some subtle details like folds and bulges, but the reconstructed geometry is inevitably bottlenecked by the capability of the model space.
Recently, the advantages of SDF-based \cite{yariv2023bakedsdf,wang2021neus,yariv2021volume} and Gaussian-based methods \cite{kerbl3Dgaussians,guedon2024sugar,Huang2DGS2024,Yu2024GOF} over traditional reconstruction methods \cite{kazhdan2006poisson,barnes2009patchmatch,furukawa2009accurate,broadhurst2001probabilistic,seitz1999photorealistic} motivate us to employ new representations for eyelid modeling. Considering the ability of both static and dynamic reconstruction, SDF is the superior choice due to its competitive performance in static geometry modeling and its proven effectiveness in dynamic scenes \cite{cai2022neural,shao2023tensor4d}, while Gaussian-based methods require further exploration in dynamic geometry reconstruction. Compared with widely used meshes, SDF can naturally avoid self-interaction and seamlessly integrate with continuous neural fields \cite{park2019deepsdf}, which is cost-effective and worth exploring in eyelid reconstruction.

In this paper, we propose a novel SDF-based method that achieves high-quality geometry reconstruction and animation of eyelids with an RGB video recorded by a single mobile phone. 
The input video contains an actor-oriented sequence with the camera moving in front and the actor performing different gaze directions, which is only a sparse sampling of view-gaze pairs and thus easy to shoot.
With such a video, we learn a dynamic neural SDF field to model the dynamic eyelids. The dynamics are modeled by a canonical hyper-space \cite{park2021hypernerf} cooperating with an invertible deformation field \cite{cai2022neural}. 
The key to our method is to consider eyeball influence on eyelids from both perspectives of eyeball shapes and motions. These can provide additional constraints over the limited inputs and help to reconstruct gaze-related details during eye movements. 
Specifically, for eyeball shapes, we propose a contact loss to make the eyelids tightly adhere to the eyeball surface, which provides geometry priors to compensate for the lack of depth information in monocular RGB recording. For eyeball motions, we learn a novel gaze-dependent adaptive anchor grid for eyelid representation. This provides a gaze-varying geometry encoding to better model the subtle details of dynamic eyelids.

To acquire eyeball information to assist eyelid reconstruction, we design an automatic eyeball calibration method to identify the 3D eyeball positions based on differentiable mesh rendering \cite{Laine2020diffrast}. 
Different from previous methods \cite{wen2020accurate,wang2017real,sun2015real,lu2020improved}, our calibration does not need 2D or 3D gaze targets as input. It works with videos containing sparse samples of view-gaze pairs, so no additional recording is required. 
By leveraging the idea that only the correct 3D eyeball position can depict the appearance of all gazes, 
we optimize the eyeball position and per-frame rotations via aligning the rendering results with the eyeball appearance (Sec. \ref{sec:eyeball_calibration}).

Apart from reconstruction, the eyeball information also benefits the eyelid animation. As the movement of eyelids often contains identity-dependent deformation, it is costly to cover all the deformation by man-made patterns \cite{wen2017real} or empirical modeling \cite{wood2016learning}. To solve this, we train an eyelid control network (Sec. \ref{sec:gaze_animation}) for each identity to model the identity-specific deformation driven by the eyeball rotation. We learn the mapping along with the dynamic neural SDF field, making it possible to control the complicated eyelid movements by simple and semantic parameters (i.e., eyeball rotations). As there are usually some eye-irrelevant motions (e.g., moving head, pursing lips) during real-data capture, we introduce a freely learnable latent code to model this part of motions and a strategy to disentangle the two kinds of motions.

In summary, our main contributions are listed as follows:
\begin{itemize}
	\item To the best of our knowledge, we are the first to achieve detailed eyelid reconstruction and animation by only using an RGB video from one mobile phone as input.
	\item We utilize eyeball information to improve eyelid reconstruction, including both eyeball shapes and motions, by a contact loss and a gaze-dependent adaptive anchor grid, respectively.
	\item We develop an eyelid control module to achieve semantic animation control over the identity-specific eyelid deformation which is well trained in a disentanglement strategy.
\end{itemize}

\section{Related Works}
\subsection{3D Head Avatars}
Early works \cite{garrido2016reconstruction,thies2016face2face} adopt the 3D morphable model \cite{blanz1999morphable} to recover the shape and appearance of human faces. Recently, many works \cite{Gafni_2021_CVPR,zielonka2023instant,xu2023avatarmav,gao2022reconstructing,zhao2023havatar,chen2023implicit} utilize the neural radiance field \cite{mildenhall2021nerf} to model the human face, while  other studies \cite{xiang2024flashavatar,ma20243d,qian2024gaussianavatars,xu2024gaussian} employ 3D Gaussian splats \cite{kerbl3Dgaussians} for the same purpose. However, these methods usually focus on the photorealistic appearance and can hardly create accurate geometry of the faces. 
For better geometry, \citet{cao2022Authentic} utilize high-quality priors to generate photorealistic avatars. Neural Head Avatars \cite{grassal2022neural} represents detailed geometry through per-vertex offsets on top of a coarse template. IMavatar \cite{zheng2022avatar} learns a canonical space and a series of expression deformations to model the dynamic geometry, while PointAvatar \cite{zheng2023pointavatar} employs point cloud representation to achieve faster rendering and better appearance. FLARE \cite{bharadwaj2023flare} apply the physically-based rendering to disentangle the material and illumination, resulting in a relightable avatar.
However, these methods mainly focus on the whole face, and the geometry of eye region is relatively coarse.

\subsection{Eye Reconstruction and Animation} The research of eye modeling mainly focus on the eyeball and eyelid. The high-quality eyeball reconstruction and rigging is well addressed by a series of work \cite{berard2014high,berard2016lightweight,berard2019practical}. For eyelid modeling, \citet{bermano2015detailed} reconstruct detailed eyelids based on a multi-camera capture setup, which is too expensive for consumer-level applications. To solve this, \citet{wood20163d} build a 3D morphable model of the eye from high-fidelity scans, which can be fitted with a single image. Similarly, \citet{wen2017real} design a set of eyelid blendshapes and propose a real-time eyelid tracking method based on one RGB-D camera, but the quality of subtle details is constrained by the capability of the model space. 
As for eyelid animation, gaze or eyeball rotation is a common parameter for controlling eyelid deformation in traditional scheme \cite{neog2016interactive,wood2016learning,kerbiriou2022detailed}. Different from theses methods, our eyelid control is based on a neural architecture like \citet{schwartz2020eyes}, \citet{li2022eyenerf}, and \citet{ShellNeRF}, but with a more lightweight input and more attention to the geometry. To the best of our knowledge, we are the first to make detailed eyelid reconstruction and animation based on lightweight captures.

\subsection{Neural Implicit Surfaces}
Different from neural radiance field \cite{mildenhall2021nerf} that models the scene by density, neural implicit surfaces \cite{yariv2023bakedsdf,wang2021neus,yariv2021volume} represent the geometry as a neural SDF field \cite{park2019deepsdf}, achieving better static surface reconstruction with multi-view RGB inputs. NDR \cite{cai2022neural} and Tensor4D \cite{shao2023tensor4d} extend the method to dynamic scenes by applying deformation field \cite{park2021hypernerf,park2021nerfies,pumarola2021d,tretschk2021non,fang2022fast}, but rely on RGB-D and multi-camera setup to guarantee the geometry quality. Despite their good results on the large-scale geometry, some subtle details are still missing. For detailed reconstruction, recent works \cite{fu2022geo,li2023neuralangelo,rosu2023permutosdf,cai2023neuda} have made explorations in static reconstruction. In this paper, we propose an gaze-dependent adaptive anchor grid to extend the static techniques to dynamic scenes, achieving detailed reconstruction of the moving eyelids.

\section{Method}
\begin{figure*}[!t]
	\centering
	\includegraphics[width=1.0\linewidth]{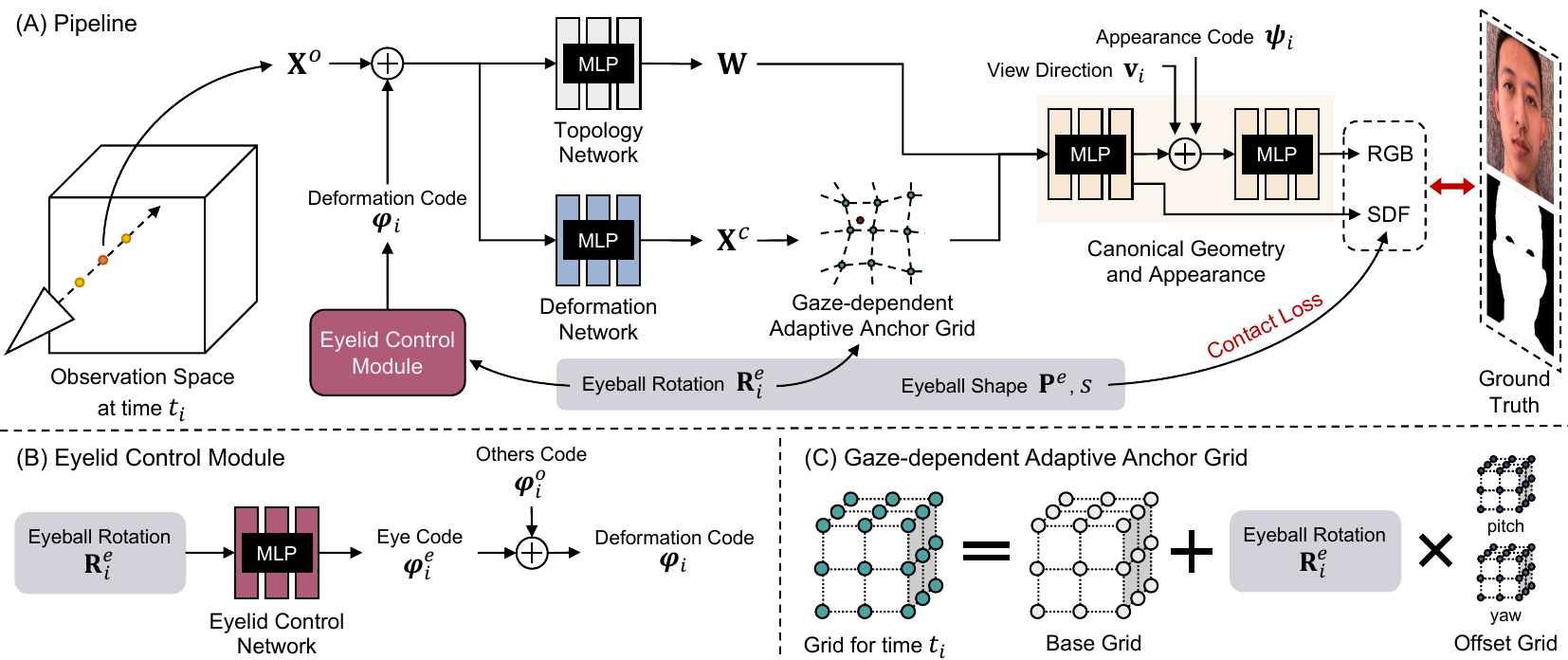}
	\caption{An overview of our method. (A) Our method models the moving eyelids as a dynamic neural SDF field, which is achieved by a canonical hyper-space with deformation. For a sampled 3D point in the observation space, the topology and deformation network convert it to the canonical hyper-space, where the SDF field is defined. Then, its point color and SDF predicted by MLPs are used to apply volume rendering for training on RGB images.  (B) This module divides the dynamic information into eye motions and others. The eye motions are modeled by the latent code mapped from eyeball rotations, and the other movements are modeled by a learnable latent code.  (C) We encode the geometry feature of a 3D canonical point by the positions of neighbor anchors. For each frame, the anchor positions are determined by a learnable base grid plus the linear combination of two learnable offset grids based on eyeball rotations.}
	\label{fig:pipeline}
\end{figure*}

The architecture of the proposed method is illustrated in Fig. \ref{fig:pipeline}. The input is an actor-oriented RGB video captured by one camera (e.g., mobile phone) moving in the front of the actor performing different gaze directions. Additionally, we adopt the video segmentation method \cite{lin2022robust} to get the mask of the actor. Our goal is to get a dynamic neural SDF field for eyelid reconstruction and animation. In order to acquire high-quality results with such limited inputs, we leverage the eyeball information (geometry and motions) to provide additional guidance.
Specifically, we obtain the masks of the eyelid and iris by the commercial service of SenseTime\footnote{\url{www.sensetime.com}} and design an eyeball calibration method to get the eyeball parameters we need. Note that the calibration suits the same capture setup and does not require additional inputs (Sec. \ref{sec:eyeball_calibration}). 
Then, we use the eyeball parameters to assist the learning of dynamic neural SDF field (Sec. \ref{sec:sdf_field}) and achieve
the semantic animation control of eyelids (Sec. \ref{sec:gaze_animation}).
The optimization details can be found in Sec. \ref{sec:opt_details}. Note that our method can process both eyes of the human face, but to simplify the explanation, most descriptions are in the context of one single eye.

\subsection{Eyeball Calibration}\label{sec:eyeball_calibration}
\begin{figure}[!t]
	\centering
	\includegraphics[width=1.0\linewidth]{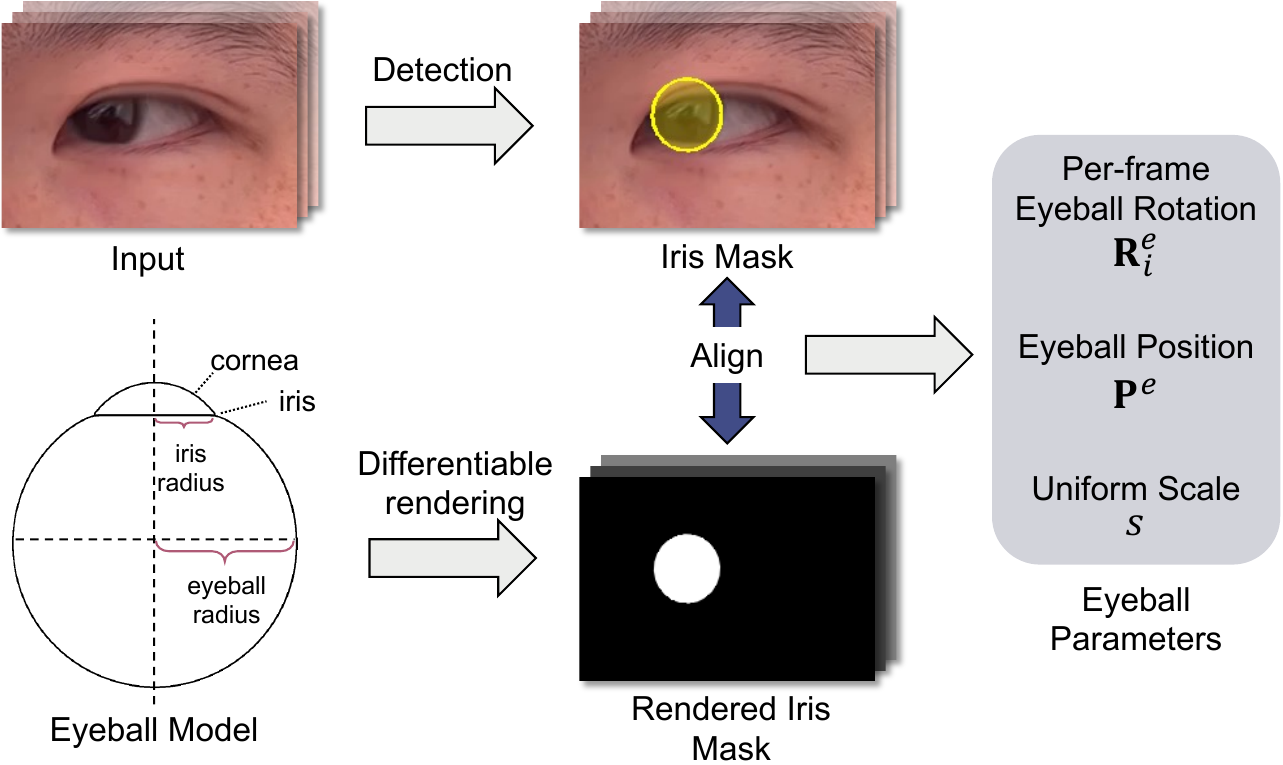}
	\caption{The eyeball calibration is based on an 3D eyeball model with physiological prior. We apply differentiable rendering to optimize eyeball parameters by aligning iris masks.}
	\label{fig:eyeball}
\end{figure}

The goal of the eyeball calibration is to get the eyeball positions and rotations for assisting the eyelid reconstruction. As human eyeballs have strong prior in both shapes and dimensions \cite{von1925helmholtz,kaufman2003adler}, we leverage these physiological priors to simplify the calibration problem. The key idea of our method is that only the eyeball with correct position can depict the appearance of all gazes. Therefore, given a 3D eyeball template $\mathcal{T}$ with physiologically appropriate ratio of the iris and eyeball radius, we optimize one eyeball position, one uniform scale, and n-frame eyeball rotations by aligning the rendering results with the appearance of each frame.
Our eyeball template is shown in Fig. \ref{fig:eyeball}. It contains a cut sphere and a convex cornea that satisfies the mathematical modeling in \cite{nishino2004world}.
As the iris is the noticeable feature of human eyes, we take the iris mask as the target of alignment, which can be described as
\begin{equation}
    E(\textbf{P}^{e},\textbf{R}^{e}_{i},s;\textbf{P}^{c}_{i},\mathcal{T},\mathcal{M}) = \sum_{i}^{n}{\|\mathcal{M}_i-\hat{\mathcal{M}}_i\|_2^2}
\end{equation}
\begin{equation}
	\hat{\mathcal{M}}_i=\Pi { (\mathcal{T},s,\textbf{P}^{e},\textbf{P}^{c}_{i},\textbf{R}^{e}_{i})}
\end{equation}
where $\textbf{P}^{e}\in \mathbb{R}^3$ is the eyeball position, and $s\in \mathbb{R}$ is the scale of eyeball template. They are shared by all frames. $n$ is the number of frames. $\textbf{P}^{c}_{i}\in SE(3)$ is the camera pose of the i-th frame. $\textbf{R}^{e}_{i}=[p_i,y_i]\in  \mathbb{R}^2$ (represented as pitch and yaw in the model space) is the eyeball rotation of the i-th frame. $\mathcal{M}$ is the ground-truth iris mask. $\hat{\mathcal{M}}$ is the rendered iris mask of the 3D eyeball template, which is got by the differentiable mesh rendering $\Pi  (\cdot)$ \cite{Laine2020diffrast}. Note that we solve the parameters of the two eyes separately.

The above optimization relies on good initial values. In order to avoid the local minimal, we have some strategies to initialize the parameters. Details can be found in the supplementary materials.

\subsection{Dynamic Neural SDF Field for Eyelid}\label{sec:sdf_field}
We model the moving eyelids (together with the other facial regions) as a dynamic neural SDF field. 
Similar to the previous works \cite{park2021nerfies,park2021hypernerf}, we construct the deformation between each observation space and the canonical space. Specifically, the dynamic modeling is achieved by a canonical hyper-space \cite{park2021hypernerf} cooperating with the invertible deformation field \cite{cai2022neural}. Further, we leverage the eyeball information to provide extra constraints on geometry and a gaze-varying encoding for subtle details.

\subsubsection{Preliminaries: NeuS for Dynamic Scenes}
NeuS \cite{wang2021neus} is an neural implicit surface reconstruction method that can reconstruct the surface of a static object by inputting posed images. It represents the geometry as the zero-level set of signed distance function (SDF) $S=\{\textbf{x}\in \mathbb{R}^3|f(\textbf{x})=0\}$, For a particular 3D point $\textbf{p}$ in the space, its SDF value and geometric feature  $\textbf{z}$ are predicted by an MLP $F_s(\textbf{p})$. To train this MLP, they propose a probability function to convert the SDF values of $N$ points sampled along a ray $r=\{\textbf{p}_{k}=\textbf{o}+t_k\textbf{v}|k=1,...,N,t_k<t_{k+1}\}$ to density value for volume rendering:
\begin{equation}
	\hat{C}=\sum_{k}^{N}T_{k}\alpha_{k}\textbf{c}_{k}, \quad T_{k}=\prod_{j=1}^{k-1}(1-\alpha_{j})
\end{equation}
Here, $\textbf{c}_k$ is the color of the k-th points, which is predicted by another trainable MLP $F_c(\textbf{p}_k,\textbf{n}_k,\textbf{v},\textbf{z}_k)$ from normal $\textbf{n}_k$, view direction $\textbf{v}$, and geometric feature $\textbf{z}_k$. The discrete opacity values $\alpha_{k}$ is calculated by
\begin{equation}
	\alpha_{k}=\max(\frac{\Phi_{s}(f(\textbf{p}(t_k)))-\Phi_{s}(f(\textbf{p}(t_{k+1})))}{\Phi_{s}(f(\textbf{p}(t_k)))}, 0)
\end{equation}
The $\Phi_{s}(x)$ is defined as $\Phi_{s}(x)=(1+e^{-sx})^{-1}$. The $s$ is a trainable parameter. The training is based on the photometric loss that minimizes the difference between the rendered images and input images.

To adapt NeuS \cite{wang2021neus} to dynamic scenes, NDR \cite{cai2022neural} construct the deformation between observation and canonical space like dynamic NeRF methods \cite{park2021nerfies,park2021hypernerf}. Further, they employ a strictly invertible bijective mapping based on the continuous homeomorphic mapping $\mathcal{D}_i:\mathbb{R}^3\to \mathbb{R}^3$. It maps the observation point $\textbf{X}^o=[x_i,y_i,z_i]$ at time $t_i$ back to the 3D canonical position $\textbf{X}^c=[x,y,z]$. As $\mathcal{D}_i$ is invertible, the deformed surface at time $t_i$ can be obtained by
\begin{equation}
	U_{i}=\{\mathcal{D}_{i}^{-1}([x,y,z])|\forall[x,y,z]\in U\}
\end{equation}
and the correspondence between deformed points can be factorized as
\begin{equation}
	[x_j,y_j,z_j]=\mathcal{D}_{j}^{-1}\circ \mathcal{D}_{i}([x_i,y_i,z_i])
\end{equation}

However, the invertible representation only effectively models the topology-preserving motions. To better model the changes in topology, \citet{cai2022neural} introduce the topology-aware design \cite{park2021hypernerf} into their deformation field. Specifically, the 3D position $\textbf{X}^o$ observed at time $t_i$ is mapped to topology coordinates by the  mapping $\mathcal{W}:\mathbb{R}^3\to \mathbb{R}^m$. Cooperating with the 3D canonical position $\textbf{X}^c$, the canonical hyper-space is represented as
\begin{equation}
	\textbf{X}=[\textbf{X}^c,\textbf{W}]
\end{equation}
where $\textbf{W}\in \mathbb{R}^m$ is the topology coordinates. In practice, the invertible deformation mapping and the topology mapping are modeled by two separate MLPs:
\begin{equation}
	\textbf{X}^c=F_d(\textbf{X}^o,\boldsymbol{\varphi}_i),\quad \textbf{W}=F_w(\textbf{X}^o,\boldsymbol{\varphi}_i)
\end{equation}
where $\boldsymbol{\varphi}_i$ is the trainable deformation code at time $t_i$. Then, the neural implicit surface is defined as the zero-level set of an SDF on the canonical hyper-space:
\begin{equation}
	S=\{\textbf{x}\in \mathbb{R}^{3+m}|f(\textbf{x})=0\}
\end{equation}

Similar to NeuS \cite{wang2021neus}, the SDF value is converted to density for training based on volume rendering. The point color at time $t_i$ is predicted from the 3D canonical position $\textbf{X}^c$, the normal $\textbf{n}^c$, the view direction $\textbf{v}^c$, and the geometric feature $\textbf{z}^{c}$ conditioned a time-varying appearance code $\boldsymbol{\psi}_i$, formulated as
\begin{equation}
	\textbf{c}_i=F_c(\textbf{X}^c,\textbf{n}^c,\textbf{v}_i^c,\textbf{z}^{c},\boldsymbol{\psi}_i)
\end{equation}

\subsubsection{Gaze-dependent Adaptive Anchor Grid} 
Despite the impressive results of NDR \cite{cai2022neural} on large-scale geometry, it still struggles to recover fine-grain geometry of dynamic scenes. To reconstruct subtle details, some solutions \cite{fu2022geo,li2023neuralangelo,rosu2023permutosdf,cai2023neuda} are proposed for static reconstruction, but few works solve it in dynamic scenes. One naive solution is to directly apply the static techniques on the 3D canonical space, but this scheme inevitably misses some time-varying information and generates low-quality results (Sec. \ref{sec:ablation}).

Instead, we develop a gaze-dependent adaptive anchor grid to improve subtle details of dynamic eyelid reconstruction. Following NeuDA \cite{cai2023neuda}, we attribute the shortage of details to the poor utilization of the spatial context. For example, NeuS \cite{wang2021neus}, as a method purely based on MLPs, implicitly utilizes the spatial context through the continuity of MLPs, which is an indirect way and often leads to over-smooth results. Voxel grid approaches \cite{liu2020neural,fridovich2022plenoxels,sun2022direct,takikawa2021neural} forward a step by explicitly encoding the geometry feature of a particular 3D point using the surrounding eight vertices, followed by the hierarchical structure \cite{muller2022instant,takikawa2021neural,wang2023neus2} for different levels of representation. However, as they uniformly split the 3D space, the distribution of vertices may not be optimal for a particular object because the distribution of object details is spatially variant, which means we need model the geometry features adaptively (e.g., dense for subtle details, sparse for large-scale geometry).
This issue may also occur when making local modeling based on facial landmarks \cite{chen2023implicit}, which are manually defined and not fully adaptive.
NeuDA \cite{cai2023neuda} solve this problem by introducing a multi-level learnable anchor grid. For a specific sample point $\textbf{p}\in \mathbb{R}^3$, the geometry feature of level $l$ is represented as the interpolated frequency embedding of the 8-nearest neighbor vertices (anchors):
\begin{equation}
	G^l(\textbf{p},\textbf{A})=\sum_{u}w_u^l\cdot \gamma(\textbf{a}_u^l), \quad u=1,...,8
\end{equation}
\begin{equation}
	\gamma(\textbf{a}^l)=(\sin(2^l \pi \textbf{a}^l),\cos(2^l\pi \textbf{a}^l))
\end{equation}
where $\textbf{A}\in \mathbb{R}^{N_a\times 3}$ denotes the anchor grid. $\textbf{a}_u^l\in \mathbb{R}^3$ is the learnable anchors of level $l$ around $\textbf{p}$, and $w_u^l$ is the trilinear interpolation weight. $\gamma(\cdot)$ is the frequency embedding proposed by \cite{mildenhall2021nerf}. Finally, the geometry encoding is represented as the concatenation of per-level frequency embeddings:
\begin{equation}
	F_g(\textbf{p},\textbf{A})=[\textbf{p},G^1(\textbf{p},\textbf{A}),G^2(\textbf{p},\textbf{A}),...,G^l(\textbf{p},\textbf{A})]
\end{equation}

\begin{figure}[!t]
	\centering
	\includegraphics[width=1.0\linewidth]{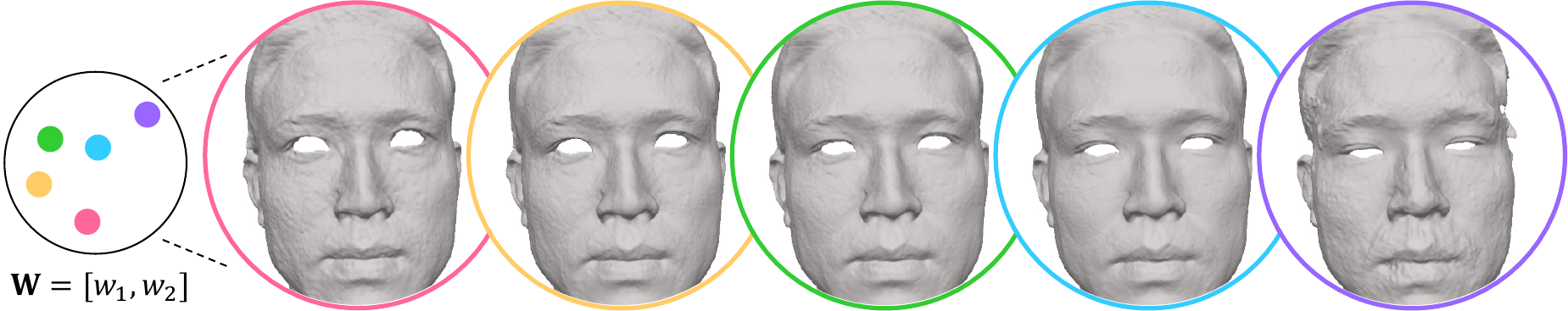}
	\caption{Topology coordinates model various shape templates of the eyelids.}
	\label{fig:wspace}
\end{figure}

Different from NeuDA \cite{cai2023neuda} that learns a fixed anchor grid, we learn a gaze-dependent anchor grid by modeling the relationship between eyeball rotations and anchor positions. As we represent the deformation field by the hyper-space (3D canonical space plus the topology coordinates), it can be taken as a set of 3D canonical templates defined by the topology coordinates. Similar to HyperNeRF \cite{park2021hypernerf}, if we visualize the hyper-space by sampling different topology coordinates (Fig. \ref{fig:wspace}), we can find the hyper-space models a family of shapes. This motivates us to learn time-varying anchor grids to represent the geometry of the moving eyelids. A direct way to achieve this is to relate the anchor grid with the topology coordinates. However, this operation will make a explosive growth in the computation graph, which brings a great burden to the device capacity. Considering the topology changes are highly relevant with the gaze poses, we associate the anchor grid with the eyeball rotations instead. 
In practice, we model the anchor grid as the linear combination of a base anchor grid $\textbf{A}_0$ and two offset grid $\textbf{A}_p, \textbf{A}_y$ that stores anchor offset related to the pitch and yaw angle, respectively. The final anchor grid of the i-th frame can be formulated as
\begin{equation}\label{eq:anchor}
	\textbf{A}_i=\textbf{A}_0+p_i\cdot \textbf{A}_p+y_i\cdot \textbf{A}_y
\end{equation}

\begin{figure}[!t]
	\centering
	\includegraphics[width=1.0\linewidth]{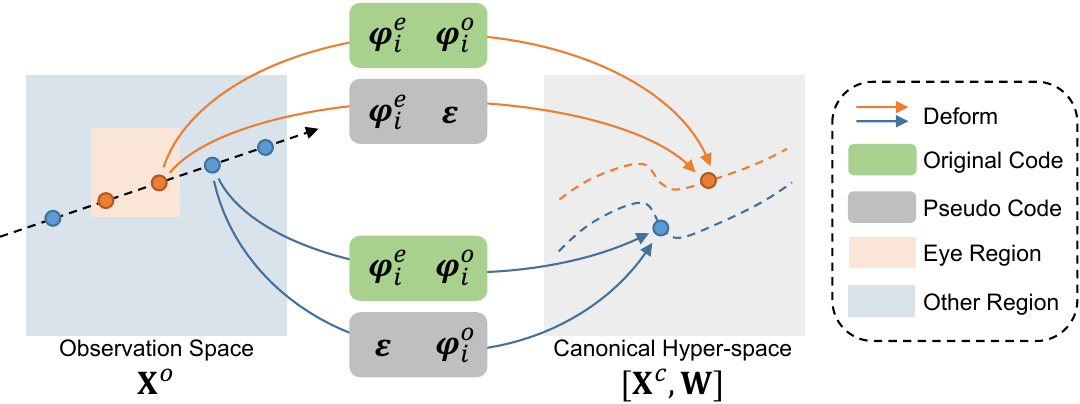}
	\caption{Illustration of disentanglement strategy. We generate pseudo deformation codes by replacing one of the parts of the original deformation code with a random latent code $\boldsymbol{\varepsilon}$. The control region of the unchanged part is encouraged to have the same hyper-space coordinates.}
	\label{fig:disentanglement}
\end{figure}

\subsection{Eyelid Control}\label{sec:gaze_animation}
In previous methods \cite{park2021hypernerf,cai2022neural}, the per-frame deformation code $\boldsymbol{\varphi}_i$ is a freely learnable latent code that cannot be controlled semantically. In order to better animate the reconstructed eyelids, we develop an eyelid control module to model the eyelid deformation as a function of the eyeball rotation.

A straightforward solution is to predict the deformation code $\boldsymbol{\varphi}_i$ from the eyeball rotation $\textbf{R}^e_i$ using an MLP, formulated as $\boldsymbol{\varphi}_i=F_e(\textbf{R}^e_i)$. However, this design assumes that the dynamic scenes only contain eye-related movements, which is almost unattainable in real-world applications because the input video inevitably contains some unconscious movements of the other facial regions.

To solve this problem, we divide the deformation into two parts (eye movements and others), and simultaneously learn a latent code $\boldsymbol{\varphi}^o_i$ to model the movements of the other facial regions. Therefore, the final deformation code is represented as
\begin{equation}
	\boldsymbol{\varphi}_i=[\boldsymbol{\varphi}^e_i,\boldsymbol{\varphi}^o_i]=[F_e(\textbf{R}^e_i),\boldsymbol{\varphi}^o_i]
\end{equation}

To ensure each part has a proper control region, we introduce a disentanglement strategy as shown in Fig. \ref{fig:disentanglement}. During training, we will generate several pseudo deformation codes by replacing one of the parts with a random latent code $\boldsymbol{\varepsilon}$, and encourage the hyper-space coordinates in the control region of the unchanged parts keep the same as before. Taking eye-region points (orange in Fig. \ref{fig:disentanglement}) as an example, although it is mapped to hyper-space by two different codes (the original code and the pseudo code with the second part changed), the two results should be the same. This can prevent the second part from controlling eye-region movements. The control region is split by the eye bounding box in the observation space, which can be automatically generated based on the eyeball parameters acquired in Sec. \ref{sec:eyeball_calibration}. Note that this strategy can be extended from \{eye, others\} to \{left eye, right eye, others\} by further splitting $\boldsymbol{\varphi}^e_i$ into two sub-codes for the left/right eye, respectively. It can also be combined with conditions to model closed eyes for generating natural eyelid movements like blinking. Details can be found in the supplementary materials.

\subsection{Optimization Details}\label{sec:opt_details}
Given an RGB video with the mask of the actor and the eyeball parameters $\{\textbf{R}^e_i,\textbf{P}^e,s\}$ acquired by eyeball calibration (Sec. \ref{sec:eyeball_calibration}), we optimize the parameters of MLPs $\{F_s,F_c,F_d,F_w,F_e\}$, the anchor grids $\{\textbf{A}_0,\textbf{A}_p,\textbf{A}_y\}$, and latent codes $\{\boldsymbol{\varphi}^o_i,\boldsymbol{\psi}_i\}$. We assume the batch size is $R$, and we sample $N$ points for each ray.

We minimize the difference between the rendered and ground-truth pixel color by
\begin{equation}
	\mathcal{L}_{color}=\frac{1}{R}\sum_{r}\|C(r)-\hat{C}(r)\|_1
\end{equation}
where $r$ is a specific ray in the volume rendering formulation \cite{mildenhall2021nerf}.

To focus on the actor, we also optimize mask loss defined as 
\begin{equation}
	\mathcal{L}_{mask}=BCE(M(r),\sum_{k}^{N}T_{r,k}\alpha_{r,k})
\end{equation}
where $BCE(\cdot)$ denotes the binary cross entropy loss, and $M(r)$ is the mask label of ray $r$.

An Eikonal loss \cite{gropp2020implicit} is introduced to regularize the signed distance function by
\begin{equation}
	\mathcal{L}_{reg}=\frac{1}{RN}\sum_{r,k}(\|\nabla f(\textbf{p}_{r,k})\|_2-1)^2
\end{equation}

Inspired by \cite{verbin2022ref}, we optimize a normal regularization loss by auxiliarily predicting a normal vector $\hat{\textbf{n}}_{r,k}$ using the MLP $F_s$. The predicted normal is encouraged to be close to the gradient of SDF:
\begin{equation}
	\mathcal{L}_{norm}=\sum_{r,k}T_{r,k}\alpha_{r,k}\|\nabla f(\textbf{p}_{r,k})-\hat{\textbf{n}}_{r,k}\|_1
\end{equation}

\begin{figure}[!t]
	\centering
	\includegraphics[width=1.0\linewidth]{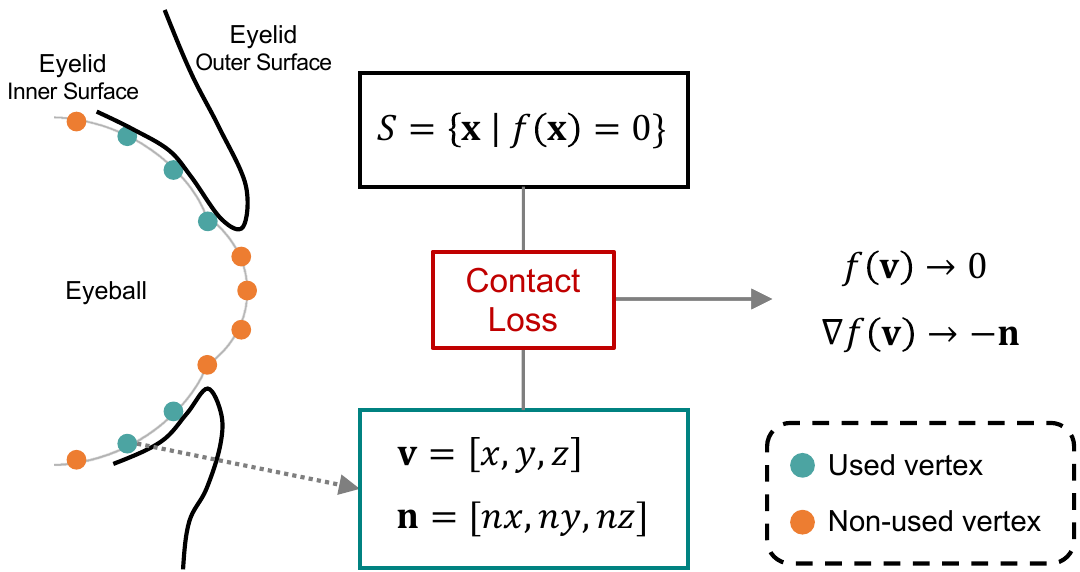}
	\caption{The contact loss encourages the inner surface of eyelid tightly adhere to the eyeball surface, which provides additional geometric constraints. The used and non-used vertices are chosen by excluding all back vertices and visible frontal vertices.}
	\label{fig:contact_loss}
\end{figure}

In addition to the above terms, we also leverage the eyeball information to provide geometry constraints for the eyelids. As shown in Fig. \ref{fig:contact_loss}, our method models both outer and inner surfaces of eyelids. We encourage the inner surface of eyelid to tightly adhere to the eyeball surface. 
In practice, we use the positions of eyeball vertices and encourage the SDF field of eyelid to have zero SDF value in these positions while holding opposite normal directions:
\begin{equation}
	\mathcal{L}_{contact}=\frac{1}{V} \sum_{i}^{V}\|f(\textbf{v}_i)\|_1+\frac{1}{V}\sum_{i}^{V}\|\nabla f(\textbf{v}_i)\cdot \textbf{n}_i+1\|_1
\end{equation}
where $\textbf{v}_i$ is the sampled position on the eyeball surface, and $\textbf{n}_i$ is the corresponding eyeball normal. $V$ is the number of samples. 
Note that all back and visible frontal vertices are excluded because they do not contact the eyelids. The former is determined by normal directions in the eyeball model space, while the latter is frame-specific and determined by whether their projected 2D positions fall within the eyelid mask.

To ensure the disentanglement strategy described in Sec. \ref{sec:gaze_animation}, we introduce disentangle loss defined as
\begin{equation}
	\begin{aligned}
		\mathcal{L}_{det}&=\frac{1}{RN}\sum_{\textbf{X}^o\in B^e}\|H(\textbf{X}^o,[\boldsymbol{\varphi}^e_i,\boldsymbol{\varepsilon}])-H(\textbf{X}^o,[\boldsymbol{\varphi}^e_i,\boldsymbol{\varphi}^o_i])\|_1\\
		&+\frac{1}{RN}\sum_{\textbf{X}^o\in B^o}\|H(\textbf{X}^o,[\boldsymbol{\varepsilon},\boldsymbol{\varphi}^o_i])-H(\textbf{X}^o,[\boldsymbol{\varphi}^e_i,\boldsymbol{\varphi}^o_i])\|_1
	\end{aligned}
\end{equation}
where $H(\textbf{X}^o,\boldsymbol{\varphi}_i)=[F_d(\textbf{X}^o,\boldsymbol{\varphi}_i),F_w(\textbf{X}^o,\boldsymbol{\varphi}_i)]$ and $\boldsymbol{\varepsilon}$ is a random latent code sampled by standard normal distribution. $B^e$ and $B^o$ are the sets of point in the eye region and the other region, respectively.

Overall, the full objective is formulated as
\begin{equation}
	\begin{aligned}
		\mathcal{L}=\mathcal{L}_{color}&+\lambda_{mask}\mathcal{L}_{mask}+\lambda_{reg}\mathcal{L}_{reg}+\lambda_{norm}\mathcal{L}_{norm}\\
		&+\lambda_{contact}\mathcal{L}_{contact}+\lambda_{det}\mathcal{L}_{det}
	\end{aligned}
\end{equation}
where $\lambda$ is the weights for different terms. In our experiments, we set $\lambda_{mask}$, $\lambda_{reg}$, $\lambda_{norm}$, $\lambda_{contact}$, and $\lambda_{det}$ as $0.1$, $0.1$, $10^{-5}$, $0.01$, and $0.1$, respectively.

\section{Experiments}
In this section, we first present the datasets and our implementation details. Then, we compare our method with the state-of-the-art methods qualitatively and quantitatively. Finally, we evaluate our key contributions via ablation study. We refer readers to our video for input examples and dynamic results. As our contributions focus on geometry reconstruction, most experiments only show geometry results. Appearance results and comparisons can be found in our supplementary materials.

\textit{Datasets.} We use a mobile phone (iPhone 13) to collect the RGB videos. In total, we recruit six participants of different ethnicities and genders. The participants are asked to perform different gaze directions while keeping the other facial parts as static as possible. The resolution of the collected data is 1080p, and the whole capture duration is about one minute. We down-sample the video in the time axis and get RGB sequences with around 500 frames. We then use Agisoft MetaShape \cite{MetaShape} to get the camera intrinsics and extrinsics. In order to evaluate the geometry results quantitatively, we additionally make four synthetic videos based on MetaHuman \cite{MetaHuman}. The videos contains 450 frames with the similar content as the real data.

\textit{Implementation Details.} The experiments are implemented with Pytorch. Similar to  \cite{atzmon2020sal}, we initialize $f(x)$ as a unit sphere. We optimize the neural networks by Adam optimizer \cite{kingma2014adam}
with a learning rate of $5\times 10^{-4}$. We sample 1,024 rays per batch, and the number of per-ray sampled points grows from 64 to 128 by 4-times up-sampling like NeuS \cite{wang2021neus}. 
The level of our adaptive anchor grid is set to 8. 
The dimensions of $\boldsymbol{\varphi}^e_i$ is 32 (16 for left and 16 for right). We learn separate $F_e$ for the left and right eye with one shared latent code $\boldsymbol{\varphi}^o_i$ of 32 dimensions. 
For coarse-to-fine training, we adopt an incremental positional encoding strategy like Nerfies \cite{park2021nerfies}. The experiments run $1.2 \times 10^{5}$ iterations for about 12 hours on one NVIDIA RTX 3090.

\begin{table}[htp]
	\footnotesize
	\caption{Quantitative comparisons on synthetic data.}
	\label{tab:geo_cmp}
	\begin{tabular}{l|c|c|c|c|c|c|c|c}
		\hline
		\multirow{2}{*}{Methods} & \multicolumn{4}{c|}{Depth Error $\downarrow$} & \multicolumn{4}{c}{Chamfer Distance $\downarrow$} \\ \cline{2-9} 
		& ID-1         & ID-2        & ID-3        & ID-4  & ID-1         & ID-2        & ID-3        & ID-4      \\
		\hline
		NDR           &     1.76       &     1.39       &     1.90     &     1.29   &     0.131       &     0.219       &     0.241       &   0.187     \\
		Tensor4D    &    1.29    &    1.11     &      1.22     &   1.39   &    0.204    &    0.226     &      0.221     &   0.201  \\
		PointAvatar    & 3.49  &   3.58    &    3.42   &   3.54       &    2.450    &    2.139   &    1.477    &    1.409    \\
		FLARE    &    1.79    &    1.81     &   1.26   &   1.48   &   0.584    &   0.527   &  0.261   &     0.382      \\
		Wen et al.    &  2.43   &  2.87  &   1.77   &    2.14   &    1.029     &  1.406    &     0.540     &    0.849      \\
		Ours          &      \textbf{0.79}     &     \textbf{0.48}     &    \textbf{0.60}     &     \textbf{0.49} &      \textbf{0.044}     &     \textbf{0.038}     &    \textbf{0.062}     &     \textbf{0.034}   \\
		\hline
	\end{tabular}
\end{table}

\subsection{Comparisons}
As we model the moving eyelid as a dynamic neural SDF field, we compare our method with the other dynamic neural SDF methods \cite{cai2022neural,shao2023tensor4d}. 
In addition, we make comparisons with the recent face-related methods based on lightweight capture \cite{zheng2023pointavatar,bharadwaj2023flare}, as well as the model-based eyelid tracking method \cite{wen2017real}.

\begin{figure*}[!t]
	\centering
	\includegraphics[width=1.0\linewidth]{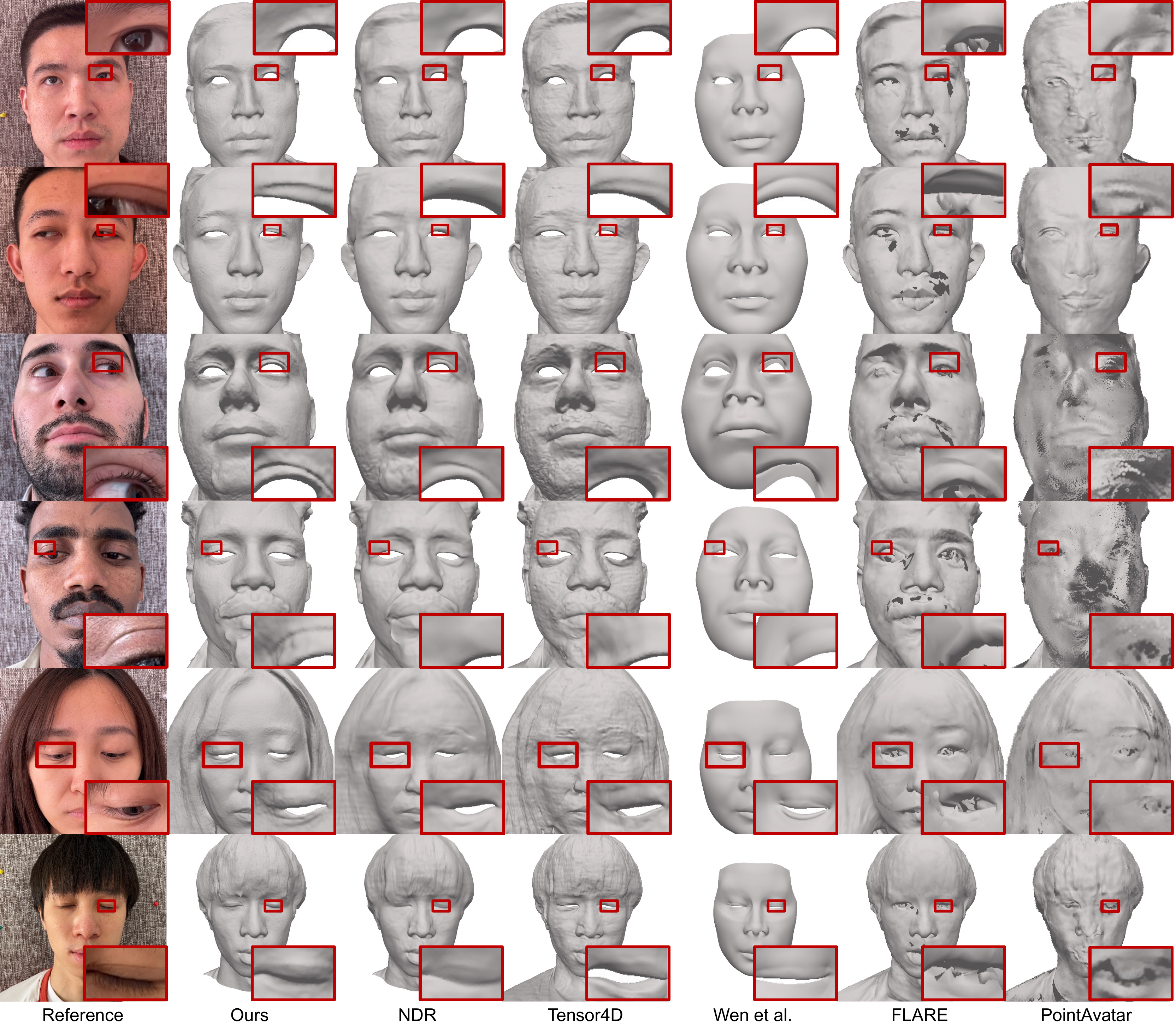}
	\caption{Qualitative comparisons on real data. The results cover different ethnicities and genders.}
	\label{fig:compare_real}
\end{figure*}

The qualitative results of real datasets are shown in Fig. \ref{fig:compare_real}. Most results are rendered with a directional light by rasterization. As PointAvatar \cite{zheng2023pointavatar} is a point-based method, we use its normal outputs to composite the shading results under the same directional light. It shows that PointAvatar \cite{zheng2023pointavatar} can only reconstruct the coarse 
contours of facial details. One possible reason is that the number of input images is one or two thousand in the original settings of PointAvatar \cite{zheng2023pointavatar}, but our input only contains hundreds of images. It may not perform well with such limited inputs. \citet{wen2017real} can generate subtle details like double-folds, but as they are model-based method, the shape of eye region is still constrained by the model capacity. FLARE \cite{bharadwaj2023flare} allows the 3D model to deform and can generate more realistic shapes, but the results suffer serious self-intersection. The results of Tensor4D \cite{shao2023tensor4d} is bumpy, while the results of NDR \cite{cai2022neural} is over-smooth. Our approach can not only reconstruct more plausible geometry but also preserve more subtle details in the eye region.

For quantitative comparisons, we use the synthetic data to calculate the geometric metrics (depth error and Chamfer distance). Specifically, for neural implicit methods (NDR \cite{cai2022neural}, Tensor4D \cite{shao2023tensor4d}, and ours), we export the mesh by marching cubes \cite{lorensen1998marching} with a resolution of 512. Then, similar to the mesh-based methods
\cite{bharadwaj2023flare,wen2017real}, the mesh is rasterized to depth map and converted to point cloud for calculating the metrics. For PointAvatar \cite{zheng2023pointavatar}, we directly use the points to calculate Chamfer distance and project the points to get a sparse depth map for calculating depth error. Note that we only calculate the metrics in the eye region, which is located by an extended bounding box of the eye mask. The average metrics of the sequence are reported in Tab. \ref{tab:geo_cmp}. Our method has a noticeable improvement compared with the previous methods.

\begin{figure}[!t]
	\centering
	\includegraphics[width=1.0\linewidth]{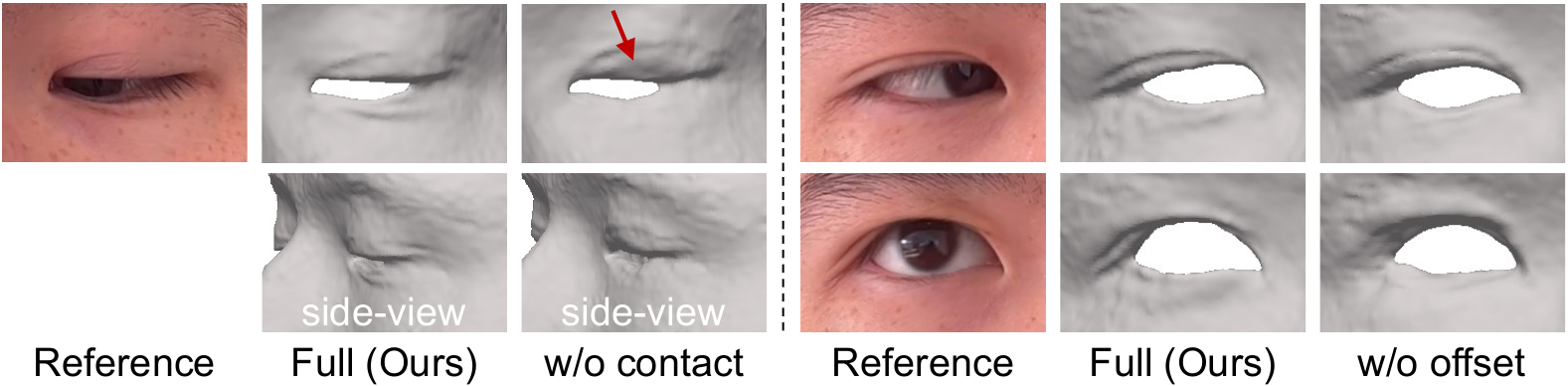}
	\caption{Ablation for reconstruction. Left: removing contact loss results in sunken shape on the upper eyelids. Right: the details becomes worse when removing the offset grids.}
	\label{fig:ablation_rec}
\end{figure}

\begin{figure}[!t]
	\centering
	\includegraphics[width=1.0\linewidth]{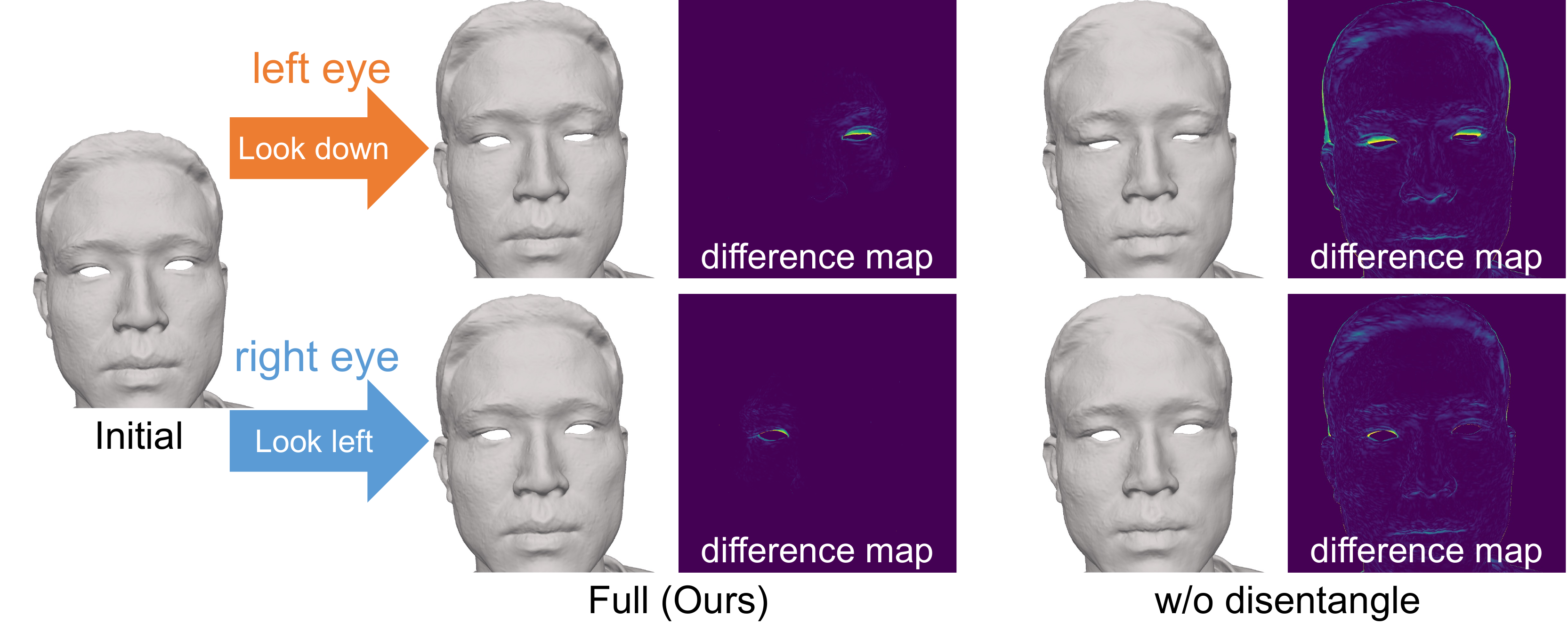}
	\caption{Ablation for animation. Without disentangle loss, the eyeball rotation cannot control the correct region of eyelids.}
	\label{fig:ablation_anim}
\end{figure}

\subsection{Ablation Study}\label{sec:ablation}
In order to evaluate our technical contributions, we design experiments to demonstrate the effectiveness of our proposed techniques for both reconstruction (contact loss and gaze-dependent adaptive anchor grid) and animation (disentanglement strategy).

For reconstruction, the first ablation (w/o contact) removes the contact loss. Results in Fig. \ref{fig:ablation_rec} show that without the contact loss, the algorithm fails to solve the depth uncertainty with such limited inputs, which leads to the sunken shape on the upper eyelid. The second ablation (w/o offset) removes the offset grid and only optimizes a fixed anchor grid. As shown in Fig. \ref{fig:ablation_rec}, it is hard for a fixed anchor grid to represent the gaze-varying shapes, which leads to the lost of the subtle details as well as the misalignment of the eyelid contour.
The quantitative results in Tab. \ref{tab:geo_ablation} also demonstrate the advantages of the proposed method. Note that we also add a third ablation (w/o both) that removes both the contact loss and the offset grid, which can be taken as a naive combination of NDR \cite{cai2022neural} and NeuDA \cite{cai2023neuda}.

\begin{table}[!t]
	\footnotesize
	\caption{Ablation study on synthetic data.}
	\label{tab:geo_ablation}
	\begin{tabular}{l|c|c|c|c|c|c|c|c}
		\hline
		\multirow{2}{*}{Methods} & \multicolumn{4}{c|}{Depth Error $\downarrow$} & \multicolumn{4}{c}{Chamfer Distance $\downarrow$} \\ \cline{2-9} 
		& ID-1         & ID-2        & ID-3        & ID-4  & ID-1         & ID-2        & ID-3        & ID-4      \\
		\hline
		Full (Ours)           &      \textbf{0.79}     &     \textbf{0.48}     &    \textbf{0.60}     &     \textbf{0.49} &      \textbf{0.044}     &     \textbf{0.038}     &    \textbf{0.062}     &     \textbf{0.034}   \\
		w/o contact    &   0.88   &    0.53   &      0.66     &   0.72   &   0.098   &    0.054     &     0.080     &   0.085  \\
		w/o offset          &     0.95    &     0.62      &    0.69     &    0.79    &      0.087    &     0.048       &      0.071      &   0.041    \\
		w/o both    &   1.00   &    0.70    &      0.90     &   0.85   &   0.122   &    0.106     &      0.113     &   0.106  \\
		\hline
	\end{tabular}
\end{table}

For animation, we compare the proposed method with the one without disentangle loss. As shown in Fig. \ref{fig:ablation_anim}, when animating the reconstructed result by changing parameters of eyeball rotations, the one without disentangle loss cannot control the eye region correctly. The difference maps clearly show that its results contain some unexpected motions like pursing lips and moving head.

\section{Discussion}
Our approach is not perfect and may miss some details like shallow creases which introduce serious ambiguity between geometry and appearance (networks can use smooth shapes with dark colors to satisfy the photometric loss). Similar issues also occur when reconstructing people with dark skins or in dark scenes. It is also hard for our method to reconstruct high-quality eyelashes, eyebrows, and bangs, which is challenging even for the high-cost method \cite{bermano2015detailed} and needs further exploration. 
The training and testing time of our method is still too long, which needs to be improved by some speed-up techniques \cite{muller2022instant,wang2023neus2} and extended by neural parameterization methods \cite{ma2022neural,lin2023single} for registration if necessary. 
Our method only considers the influence of eyeballs towards eyelids but neglects the reverse influence of eyelids towards eyeballs. It may be beneficial to optimize the eyelid and eyeball parameters jointly in the future.

\section{Conclusion}
In this paper, we propose a novel method that can make high-quality eyelid reconstruction and animation by only using lightweight captures. 
To achieve this, we leverage eyeball information to compensate for the limited inputs from both perspectives of eyeball shapes and motions. For animation, we model the eyelid deformation as a function of the eyeball rotation and introduce a disentanglement strategy to ensure the correct semantic control. Besides, we present a lightweight eyeball calibration method to acquire the eyeball parameters automatically, which does not need additional inputs. To our best knowledge, we present the first method for high-quality eyelid reconstruction and animation form lightweight inputs. We believe this technique could reduce the cost and broaden the application scenarios of realistic digital humans.

\begin{acks}
    This work was supported by the National Key R\&D Program of China (2023YFC3305600), the NSFC (No. 61822111, 62021002), the Zhejiang Provincial Natural Science Foundation (No. LDT23F02024F\\02), and the Key Research and Development Project of Tibet Autonomous Region (XZ202101ZY0019G). This work was also supported by THUIBCS, Tsinghua University, and BLBCI, Beijing Municipal Education Commission. Feng Xu is the corresponding author.
\end{acks}

\bibliographystyle{ACM-Reference-Format}
\bibliography{sample-base}


\begin{thebibliography}{80}


\ifx \showCODEN    \undefined \def \showCODEN     #1{\unskip}     \fi
\ifx \showDOI      \undefined \def \showDOI       #1{#1}\fi
\ifx \showISBNx    \undefined \def \showISBNx     #1{\unskip}     \fi
\ifx \showISBNxiii \undefined \def \showISBNxiii  #1{\unskip}     \fi
\ifx \showISSN     \undefined \def \showISSN      #1{\unskip}     \fi
\ifx \showLCCN     \undefined \def \showLCCN      #1{\unskip}     \fi
\ifx \shownote     \undefined \def \shownote      #1{#1}          \fi
\ifx \showarticletitle \undefined \def \showarticletitle #1{#1}   \fi
\ifx \showURL      \undefined \def \showURL       {\relax}        \fi
\providecommand\bibfield[2]{#2}
\providecommand\bibinfo[2]{#2}
\providecommand\natexlab[1]{#1}
\providecommand\showeprint[2][]{arXiv:#2}

\bibitem[Agisoft(2023)]%
        {MetaShape}
\bibfield{author}{\bibinfo{person}{Agisoft}.} \bibinfo{year}{2023}\natexlab{}.
\newblock \bibinfo{booktitle}{\emph{Agisoft Metashape Professional}}.
\newblock


\bibitem[Atzmon and Lipman(2020)]%
        {atzmon2020sal}
\bibfield{author}{\bibinfo{person}{Matan Atzmon} {and} \bibinfo{person}{Yaron
  Lipman}.} \bibinfo{year}{2020}\natexlab{}.
\newblock \showarticletitle{Sal: Sign agnostic learning of shapes from raw
  data}. In \bibinfo{booktitle}{\emph{Proceedings of the IEEE/CVF Conference on
  Computer Vision and Pattern Recognition}}. \bibinfo{pages}{2565--2574}.
\newblock


\bibitem[Barnes et~al\mbox{.}(2009)]%
        {barnes2009patchmatch}
\bibfield{author}{\bibinfo{person}{Connelly Barnes}, \bibinfo{person}{Eli
  Shechtman}, \bibinfo{person}{Adam Finkelstein}, {and} \bibinfo{person}{Dan~B
  Goldman}.} \bibinfo{year}{2009}\natexlab{}.
\newblock \showarticletitle{PatchMatch: A randomized correspondence algorithm
  for structural image editing}.
\newblock \bibinfo{journal}{\emph{ACM Trans. Graph.}} \bibinfo{volume}{28},
  \bibinfo{number}{3} (\bibinfo{year}{2009}), \bibinfo{pages}{24}.
\newblock


\bibitem[B{\'e}rard et~al\mbox{.}(2016)]%
        {berard2016lightweight}
\bibfield{author}{\bibinfo{person}{Pascal B{\'e}rard}, \bibinfo{person}{Derek
  Bradley}, \bibinfo{person}{Markus Gross}, {and} \bibinfo{person}{Thabo
  Beeler}.} \bibinfo{year}{2016}\natexlab{}.
\newblock \showarticletitle{Lightweight eye capture using a parametric model}.
\newblock \bibinfo{journal}{\emph{ACM Transactions on Graphics (TOG)}}
  \bibinfo{volume}{35}, \bibinfo{number}{4} (\bibinfo{year}{2016}),
  \bibinfo{pages}{1--12}.
\newblock


\bibitem[B{\'e}rard et~al\mbox{.}(2019)]%
        {berard2019practical}
\bibfield{author}{\bibinfo{person}{Pascal B{\'e}rard}, \bibinfo{person}{Derek
  Bradley}, \bibinfo{person}{Markus Gross}, {and} \bibinfo{person}{Thabo
  Beeler}.} \bibinfo{year}{2019}\natexlab{}.
\newblock \showarticletitle{Practical Person-Specific Eye Rigging}. In
  \bibinfo{booktitle}{\emph{Computer Graphics Forum}},
  Vol.~\bibinfo{volume}{38}. Wiley Online Library, \bibinfo{pages}{441--454}.
\newblock


\bibitem[B{\'e}rard et~al\mbox{.}(2014)]%
        {berard2014high}
\bibfield{author}{\bibinfo{person}{Pascal B{\'e}rard}, \bibinfo{person}{Derek
  Bradley}, \bibinfo{person}{Maurizio Nitti}, \bibinfo{person}{Thabo Beeler},
  {and} \bibinfo{person}{Markus~H Gross}.} \bibinfo{year}{2014}\natexlab{}.
\newblock \showarticletitle{High-quality capture of eyes.}
\newblock \bibinfo{journal}{\emph{ACM Trans. Graph.}} \bibinfo{volume}{33},
  \bibinfo{number}{6} (\bibinfo{year}{2014}), \bibinfo{pages}{223--1}.
\newblock


\bibitem[Bermano et~al\mbox{.}(2015)]%
        {bermano2015detailed}
\bibfield{author}{\bibinfo{person}{Amit Bermano}, \bibinfo{person}{Thabo
  Beeler}, \bibinfo{person}{Yeara Kozlov}, \bibinfo{person}{Derek Bradley},
  \bibinfo{person}{Bernd Bickel}, {and} \bibinfo{person}{Markus Gross}.}
  \bibinfo{year}{2015}\natexlab{}.
\newblock \showarticletitle{Detailed spatio-temporal reconstruction of
  eyelids}.
\newblock \bibinfo{journal}{\emph{ACM Transactions on Graphics (TOG)}}
  \bibinfo{volume}{34}, \bibinfo{number}{4} (\bibinfo{year}{2015}),
  \bibinfo{pages}{1--11}.
\newblock


\bibitem[Bharadwaj et~al\mbox{.}(2023)]%
        {bharadwaj2023flare}
\bibfield{author}{\bibinfo{person}{Shrisha Bharadwaj}, \bibinfo{person}{Yufeng
  Zheng}, \bibinfo{person}{Otmar Hilliges}, \bibinfo{person}{Michael~J Black},
  {and} \bibinfo{person}{Victoria~Fernandez Abrevaya}.}
  \bibinfo{year}{2023}\natexlab{}.
\newblock \showarticletitle{FLARE: Fast Learning of Animatable and Relightable
  Mesh Avatars}.
\newblock \bibinfo{journal}{\emph{ACM Transactions on Graphics (TOG)}}
  \bibinfo{volume}{42}, \bibinfo{number}{6} (\bibinfo{year}{2023}),
  \bibinfo{pages}{1--15}.
\newblock


\bibitem[Blanz and Vetter(1999)]%
        {blanz1999morphable}
\bibfield{author}{\bibinfo{person}{V Blanz} {and} \bibinfo{person}{T Vetter}.}
  \bibinfo{year}{1999}\natexlab{}.
\newblock \showarticletitle{A Morphable Model for the Synthesis of 3D Faces}.
  In \bibinfo{booktitle}{\emph{26th Annual Conference on Computer Graphics and
  Interactive Techniques (SIGGRAPH 1999)}}. ACM Press,
  \bibinfo{pages}{187--194}.
\newblock


\bibitem[Broadhurst et~al\mbox{.}(2001)]%
        {broadhurst2001probabilistic}
\bibfield{author}{\bibinfo{person}{Adrian Broadhurst}, \bibinfo{person}{Tom~W
  Drummond}, {and} \bibinfo{person}{Roberto Cipolla}.}
  \bibinfo{year}{2001}\natexlab{}.
\newblock \showarticletitle{A probabilistic framework for space carving}. In
  \bibinfo{booktitle}{\emph{Proceedings eighth IEEE international conference on
  computer vision. ICCV 2001}}, Vol.~\bibinfo{volume}{1}. IEEE,
  \bibinfo{pages}{388--393}.
\newblock


\bibitem[Cai et~al\mbox{.}(2023)]%
        {cai2023neuda}
\bibfield{author}{\bibinfo{person}{Bowen Cai}, \bibinfo{person}{Jinchi Huang},
  \bibinfo{person}{Rongfei Jia}, \bibinfo{person}{Chengfei Lv}, {and}
  \bibinfo{person}{Huan Fu}.} \bibinfo{year}{2023}\natexlab{}.
\newblock \showarticletitle{NeuDA: Neural Deformable Anchor for High-Fidelity
  Implicit Surface Reconstruction}. In \bibinfo{booktitle}{\emph{Proceedings of
  the IEEE/CVF Conference on Computer Vision and Pattern Recognition}}.
  \bibinfo{pages}{8476--8485}.
\newblock


\bibitem[Cai et~al\mbox{.}(2022)]%
        {cai2022neural}
\bibfield{author}{\bibinfo{person}{Hongrui Cai}, \bibinfo{person}{Wanquan
  Feng}, \bibinfo{person}{Xuetao Feng}, \bibinfo{person}{Yan Wang}, {and}
  \bibinfo{person}{Juyong Zhang}.} \bibinfo{year}{2022}\natexlab{}.
\newblock \showarticletitle{Neural surface reconstruction of dynamic scenes
  with monocular rgb-d camera}.
\newblock \bibinfo{journal}{\emph{Advances in Neural Information Processing
  Systems}}  \bibinfo{volume}{35} (\bibinfo{year}{2022}),
  \bibinfo{pages}{967--981}.
\newblock


\bibitem[Cao et~al\mbox{.}(2022)]%
        {cao2022Authentic}
\bibfield{author}{\bibinfo{person}{Chen Cao}, \bibinfo{person}{Tomas Simon},
  \bibinfo{person}{Jin~Kyu Kim}, \bibinfo{person}{Gabe Schwartz},
  \bibinfo{person}{Michael Zollhoefer}, \bibinfo{person}{Shun-Suke Saito},
  \bibinfo{person}{Stephen Lombardi}, \bibinfo{person}{Shih-En Wei},
  \bibinfo{person}{Danielle Belko}, \bibinfo{person}{Shoou-I Yu},
  \bibinfo{person}{Yaser Sheikh}, {and} \bibinfo{person}{Jason Saragih}.}
  \bibinfo{year}{2022}\natexlab{}.
\newblock \showarticletitle{Authentic volumetric avatars from a phone scan}.
\newblock \bibinfo{journal}{\emph{ACM Trans. Graph.}} \bibinfo{volume}{41},
  \bibinfo{number}{4}, Article \bibinfo{articleno}{163} (\bibinfo{date}{jul}
  \bibinfo{year}{2022}), \bibinfo{numpages}{19}~pages.
\newblock
\showISSN{0730-0301}
\urldef\tempurl%
\url{https://doi.org/10.1145/3528223.3530143}
\showDOI{\tempurl}


\bibitem[Chen et~al\mbox{.}(2023)]%
        {chen2023implicit}
\bibfield{author}{\bibinfo{person}{Chuhan Chen}, \bibinfo{person}{Matthew
  O’Toole}, \bibinfo{person}{Gaurav Bharaj}, {and} \bibinfo{person}{Pablo
  Garrido}.} \bibinfo{year}{2023}\natexlab{}.
\newblock \showarticletitle{Implicit neural head synthesis via controllable
  local deformation fields}. In \bibinfo{booktitle}{\emph{Proceedings of the
  IEEE/CVF Conference on Computer Vision and Pattern Recognition}}.
  \bibinfo{pages}{416--426}.
\newblock


\bibitem[Fang et~al\mbox{.}(2022)]%
        {fang2022fast}
\bibfield{author}{\bibinfo{person}{Jiemin Fang}, \bibinfo{person}{Taoran Yi},
  \bibinfo{person}{Xinggang Wang}, \bibinfo{person}{Lingxi Xie},
  \bibinfo{person}{Xiaopeng Zhang}, \bibinfo{person}{Wenyu Liu},
  \bibinfo{person}{Matthias Nie{\ss}ner}, {and} \bibinfo{person}{Qi Tian}.}
  \bibinfo{year}{2022}\natexlab{}.
\newblock \showarticletitle{Fast dynamic radiance fields with time-aware neural
  voxels}. In \bibinfo{booktitle}{\emph{SIGGRAPH Asia 2022 Conference Papers}}.
  \bibinfo{pages}{1--9}.
\newblock


\bibitem[Fridovich-Keil et~al\mbox{.}(2022)]%
        {fridovich2022plenoxels}
\bibfield{author}{\bibinfo{person}{Sara Fridovich-Keil}, \bibinfo{person}{Alex
  Yu}, \bibinfo{person}{Matthew Tancik}, \bibinfo{person}{Qinhong Chen},
  \bibinfo{person}{Benjamin Recht}, {and} \bibinfo{person}{Angjoo Kanazawa}.}
  \bibinfo{year}{2022}\natexlab{}.
\newblock \showarticletitle{Plenoxels: Radiance fields without neural
  networks}. In \bibinfo{booktitle}{\emph{Proceedings of the IEEE/CVF
  Conference on Computer Vision and Pattern Recognition}}.
  \bibinfo{pages}{5501--5510}.
\newblock


\bibitem[Fu et~al\mbox{.}(2022)]%
        {fu2022geo}
\bibfield{author}{\bibinfo{person}{Qiancheng Fu}, \bibinfo{person}{Qingshan
  Xu}, \bibinfo{person}{Yew~Soon Ong}, {and} \bibinfo{person}{Wenbing Tao}.}
  \bibinfo{year}{2022}\natexlab{}.
\newblock \showarticletitle{Geo-neus: Geometry-consistent neural implicit
  surfaces learning for multi-view reconstruction}.
\newblock \bibinfo{journal}{\emph{Advances in Neural Information Processing
  Systems}}  \bibinfo{volume}{35} (\bibinfo{year}{2022}),
  \bibinfo{pages}{3403--3416}.
\newblock


\bibitem[Furukawa and Ponce(2009)]%
        {furukawa2009accurate}
\bibfield{author}{\bibinfo{person}{Yasutaka Furukawa} {and}
  \bibinfo{person}{Jean Ponce}.} \bibinfo{year}{2009}\natexlab{}.
\newblock \showarticletitle{Accurate, dense, and robust multiview stereopsis}.
\newblock \bibinfo{journal}{\emph{IEEE transactions on pattern analysis and
  machine intelligence}} \bibinfo{volume}{32}, \bibinfo{number}{8}
  (\bibinfo{year}{2009}), \bibinfo{pages}{1362--1376}.
\newblock


\bibitem[Gafni et~al\mbox{.}(2021)]%
        {Gafni_2021_CVPR}
\bibfield{author}{\bibinfo{person}{Guy Gafni}, \bibinfo{person}{Justus Thies},
  \bibinfo{person}{Michael Zollh{\"o}fer}, {and} \bibinfo{person}{Matthias
  Nie{\ss}ner}.} \bibinfo{year}{2021}\natexlab{}.
\newblock \showarticletitle{Dynamic Neural Radiance Fields for Monocular 4D
  Facial Avatar Reconstruction}. In \bibinfo{booktitle}{\emph{Proceedings of
  the IEEE/CVF Conference on Computer Vision and Pattern Recognition (CVPR)}}.
  \bibinfo{pages}{8649--8658}.
\newblock


\bibitem[Games(2023)]%
        {MetaHuman}
\bibfield{author}{\bibinfo{person}{Epic Games}.}
  \bibinfo{year}{2023}\natexlab{}.
\newblock \bibinfo{booktitle}{\emph{Metahuman creator}}.
\newblock


\bibitem[Gao et~al\mbox{.}(2022)]%
        {gao2022reconstructing}
\bibfield{author}{\bibinfo{person}{Xuan Gao}, \bibinfo{person}{Chenglai Zhong},
  \bibinfo{person}{Jun Xiang}, \bibinfo{person}{Yang Hong},
  \bibinfo{person}{Yudong Guo}, {and} \bibinfo{person}{Juyong Zhang}.}
  \bibinfo{year}{2022}\natexlab{}.
\newblock \showarticletitle{Reconstructing personalized semantic facial nerf
  models from monocular video}.
\newblock \bibinfo{journal}{\emph{ACM Transactions on Graphics (TOG)}}
  \bibinfo{volume}{41}, \bibinfo{number}{6} (\bibinfo{year}{2022}),
  \bibinfo{pages}{1--12}.
\newblock


\bibitem[Garrido et~al\mbox{.}(2016)]%
        {garrido2016reconstruction}
\bibfield{author}{\bibinfo{person}{Pablo Garrido}, \bibinfo{person}{Michael
  Zollh{\"o}fer}, \bibinfo{person}{Dan Casas}, \bibinfo{person}{Levi
  Valgaerts}, \bibinfo{person}{Kiran Varanasi}, \bibinfo{person}{Patrick
  P{\'e}rez}, {and} \bibinfo{person}{Christian Theobalt}.}
  \bibinfo{year}{2016}\natexlab{}.
\newblock \showarticletitle{Reconstruction of personalized 3D face rigs from
  monocular video}.
\newblock \bibinfo{journal}{\emph{ACM Transactions on Graphics (TOG)}}
  \bibinfo{volume}{35}, \bibinfo{number}{3} (\bibinfo{year}{2016}),
  \bibinfo{pages}{1--15}.
\newblock


\bibitem[Grassal et~al\mbox{.}(2022)]%
        {grassal2022neural}
\bibfield{author}{\bibinfo{person}{Philip-William Grassal},
  \bibinfo{person}{Malte Prinzler}, \bibinfo{person}{Titus Leistner},
  \bibinfo{person}{Carsten Rother}, \bibinfo{person}{Matthias Nie{\ss}ner},
  {and} \bibinfo{person}{Justus Thies}.} \bibinfo{year}{2022}\natexlab{}.
\newblock \showarticletitle{Neural head avatars from monocular rgb videos}. In
  \bibinfo{booktitle}{\emph{Proceedings of the IEEE/CVF Conference on Computer
  Vision and Pattern Recognition}}. \bibinfo{pages}{18653--18664}.
\newblock


\bibitem[Gropp et~al\mbox{.}(2020)]%
        {gropp2020implicit}
\bibfield{author}{\bibinfo{person}{Amos Gropp}, \bibinfo{person}{Lior Yariv},
  \bibinfo{person}{Niv Haim}, \bibinfo{person}{Matan Atzmon}, {and}
  \bibinfo{person}{Yaron Lipman}.} \bibinfo{year}{2020}\natexlab{}.
\newblock \showarticletitle{Implicit geometric regularization for learning
  shapes}.
\newblock \bibinfo{journal}{\emph{arXiv preprint arXiv:2002.10099}}
  (\bibinfo{year}{2020}).
\newblock


\bibitem[Gu{\'e}don and Lepetit(2024)]%
        {guedon2024sugar}
\bibfield{author}{\bibinfo{person}{Antoine Gu{\'e}don} {and}
  \bibinfo{person}{Vincent Lepetit}.} \bibinfo{year}{2024}\natexlab{}.
\newblock \showarticletitle{SuGaR: Surface-Aligned Gaussian Splatting for
  Efficient 3D Mesh Reconstruction and High-Quality Mesh Rendering}.
\newblock \bibinfo{journal}{\emph{CVPR}} (\bibinfo{year}{2024}).
\newblock


\bibitem[Huang et~al\mbox{.}(2024)]%
        {Huang2DGS2024}
\bibfield{author}{\bibinfo{person}{Binbin Huang}, \bibinfo{person}{Zehao Yu},
  \bibinfo{person}{Anpei Chen}, \bibinfo{person}{Andreas Geiger}, {and}
  \bibinfo{person}{Shenghua Gao}.} \bibinfo{year}{2024}\natexlab{}.
\newblock \showarticletitle{2D Gaussian Splatting for Geometrically Accurate
  Radiance Fields}. In \bibinfo{booktitle}{\emph{SIGGRAPH 2024 Conference
  Papers}}. \bibinfo{publisher}{Association for Computing Machinery}.
\newblock
\urldef\tempurl%
\url{https://doi.org/10.1145/3641519.3657428}
\showDOI{\tempurl}


\bibitem[Kaufman et~al\mbox{.}(2003)]%
        {kaufman2003adler}
\bibfield{author}{\bibinfo{person}{P.L. Kaufman}, \bibinfo{person}{A. Alm},
  {and} \bibinfo{person}{F.H. Adler}.} \bibinfo{year}{2003}\natexlab{}.
\newblock \bibinfo{booktitle}{\emph{Adler's Physiology of the Eye: Clinical
  Application}}.
\newblock \bibinfo{publisher}{Mosby}.
\newblock
\showISBNx{9780323011365}
\showLCCN{86016180}
\urldef\tempurl%
\url{https://books.google.com.hk/books?id=2YlsAAAAMAAJ}
\showURL{%
\tempurl}


\bibitem[Kazhdan et~al\mbox{.}(2006)]%
        {kazhdan2006poisson}
\bibfield{author}{\bibinfo{person}{Michael Kazhdan}, \bibinfo{person}{Matthew
  Bolitho}, {and} \bibinfo{person}{Hugues Hoppe}.}
  \bibinfo{year}{2006}\natexlab{}.
\newblock \showarticletitle{Poisson surface reconstruction}. In
  \bibinfo{booktitle}{\emph{Proceedings of the fourth Eurographics symposium on
  Geometry processing}}, Vol.~\bibinfo{volume}{7}. \bibinfo{pages}{0}.
\newblock


\bibitem[Kerbiriou et~al\mbox{.}(2022)]%
        {kerbiriou2022detailed}
\bibfield{author}{\bibinfo{person}{Glenn Kerbiriou}, \bibinfo{person}{Quentin
  Avril}, \bibinfo{person}{Fabien Danieau}, {and} \bibinfo{person}{Maud
  Marchal}.} \bibinfo{year}{2022}\natexlab{}.
\newblock \showarticletitle{Detailed Eye Region Capture and Animation}. In
  \bibinfo{booktitle}{\emph{Computer Graphics Forum}},
  Vol.~\bibinfo{volume}{41}. Wiley Online Library, \bibinfo{pages}{279--282}.
\newblock


\bibitem[Kerbl et~al\mbox{.}(2023)]%
        {kerbl3Dgaussians}
\bibfield{author}{\bibinfo{person}{Bernhard Kerbl}, \bibinfo{person}{Georgios
  Kopanas}, \bibinfo{person}{Thomas Leimk{\"u}hler}, {and}
  \bibinfo{person}{George Drettakis}.} \bibinfo{year}{2023}\natexlab{}.
\newblock \showarticletitle{3D Gaussian Splatting for Real-Time Radiance Field
  Rendering}.
\newblock \bibinfo{journal}{\emph{ACM Transactions on Graphics}}
  \bibinfo{volume}{42}, \bibinfo{number}{4} (\bibinfo{date}{July}
  \bibinfo{year}{2023}).
\newblock


\bibitem[Kingma and Ba(2014)]%
        {kingma2014adam}
\bibfield{author}{\bibinfo{person}{Diederik~P Kingma} {and}
  \bibinfo{person}{Jimmy Ba}.} \bibinfo{year}{2014}\natexlab{}.
\newblock \showarticletitle{Adam: A method for stochastic optimization}.
\newblock \bibinfo{journal}{\emph{arXiv preprint arXiv:1412.6980}}
  (\bibinfo{year}{2014}).
\newblock


\bibitem[Kirschstein et~al\mbox{.}(2023)]%
        {kirschstein2023nersemble}
\bibfield{author}{\bibinfo{person}{Tobias Kirschstein},
  \bibinfo{person}{Shenhan Qian}, \bibinfo{person}{Simon Giebenhain},
  \bibinfo{person}{Tim Walter}, {and} \bibinfo{person}{Matthias Nie\ss{}ner}.}
  \bibinfo{year}{2023}\natexlab{}.
\newblock \showarticletitle{NeRSemble: Multi-View Radiance Field Reconstruction
  of Human Heads}.
\newblock \bibinfo{journal}{\emph{ACM Trans. Graph.}} \bibinfo{volume}{42},
  \bibinfo{number}{4}, Article \bibinfo{articleno}{161} (\bibinfo{date}{jul}
  \bibinfo{year}{2023}), \bibinfo{numpages}{14}~pages.
\newblock
\showISSN{0730-0301}
\urldef\tempurl%
\url{https://doi.org/10.1145/3592455}
\showDOI{\tempurl}


\bibitem[Laine et~al\mbox{.}(2020)]%
        {Laine2020diffrast}
\bibfield{author}{\bibinfo{person}{Samuli Laine}, \bibinfo{person}{Janne
  Hellsten}, \bibinfo{person}{Tero Karras}, \bibinfo{person}{Yeongho Seol},
  \bibinfo{person}{Jaakko Lehtinen}, {and} \bibinfo{person}{Timo Aila}.}
  \bibinfo{year}{2020}\natexlab{}.
\newblock \showarticletitle{Modular Primitives for High-Performance
  Differentiable Rendering}.
\newblock \bibinfo{journal}{\emph{ACM Transactions on Graphics}}
  \bibinfo{volume}{39}, \bibinfo{number}{6} (\bibinfo{year}{2020}).
\newblock


\bibitem[Li et~al\mbox{.}(2022)]%
        {li2022eyenerf}
\bibfield{author}{\bibinfo{person}{Gengyan Li}, \bibinfo{person}{Abhimitra
  Meka}, \bibinfo{person}{Franziska Mueller}, \bibinfo{person}{Marcel~C
  Buehler}, \bibinfo{person}{Otmar Hilliges}, {and} \bibinfo{person}{Thabo
  Beeler}.} \bibinfo{year}{2022}\natexlab{}.
\newblock \showarticletitle{EyeNeRF: a hybrid representation for photorealistic
  synthesis, animation and relighting of human eyes}.
\newblock \bibinfo{journal}{\emph{ACM Transactions on Graphics (TOG)}}
  \bibinfo{volume}{41}, \bibinfo{number}{4} (\bibinfo{year}{2022}),
  \bibinfo{pages}{1--16}.
\newblock


\bibitem[Li et~al\mbox{.}(2024)]%
        {ShellNeRF}
\bibfield{author}{\bibinfo{person}{Gengyan Li}, \bibinfo{person}{Kripasindhu
  Sarkar}, \bibinfo{person}{Abhimitra Meka}, \bibinfo{person}{Marcel Buehler},
  \bibinfo{person}{Franziska Mueller}, \bibinfo{person}{Paulo Gotardo},
  \bibinfo{person}{Otmar Hilliges}, {and} \bibinfo{person}{Thabo Beeler}.}
  \bibinfo{year}{2024}\natexlab{}.
\newblock \showarticletitle{{ShellNeRF: Learning a Controllable High-resolution
  Model of the Eye and Periocular Region}}.
\newblock \bibinfo{journal}{\emph{Computer Graphics Forum}}
  (\bibinfo{year}{2024}).
\newblock
\showISSN{1467-8659}
\urldef\tempurl%
\url{https://doi.org/10.1111/cgf.15041}
\showDOI{\tempurl}


\bibitem[Li et~al\mbox{.}(2023)]%
        {li2023neuralangelo}
\bibfield{author}{\bibinfo{person}{Zhaoshuo Li}, \bibinfo{person}{Thomas
  M{\"u}ller}, \bibinfo{person}{Alex Evans}, \bibinfo{person}{Russell~H
  Taylor}, \bibinfo{person}{Mathias Unberath}, \bibinfo{person}{Ming-Yu Liu},
  {and} \bibinfo{person}{Chen-Hsuan Lin}.} \bibinfo{year}{2023}\natexlab{}.
\newblock \showarticletitle{Neuralangelo: High-Fidelity Neural Surface
  Reconstruction}. In \bibinfo{booktitle}{\emph{Proceedings of the IEEE/CVF
  Conference on Computer Vision and Pattern Recognition}}.
  \bibinfo{pages}{8456--8465}.
\newblock


\bibitem[Lin et~al\mbox{.}(2023)]%
        {lin2023single}
\bibfield{author}{\bibinfo{person}{Connor Lin}, \bibinfo{person}{Koki Nagano},
  \bibinfo{person}{Jan Kautz}, \bibinfo{person}{Eric Chan},
  \bibinfo{person}{Umar Iqbal}, \bibinfo{person}{Leonidas Guibas},
  \bibinfo{person}{Gordon Wetzstein}, {and} \bibinfo{person}{Sameh Khamis}.}
  \bibinfo{year}{2023}\natexlab{}.
\newblock \showarticletitle{Single-shot implicit morphable faces with
  consistent texture parameterization}. In \bibinfo{booktitle}{\emph{ACM
  SIGGRAPH 2023 Conference Proceedings}}. \bibinfo{pages}{1--12}.
\newblock


\bibitem[Lin et~al\mbox{.}(2022)]%
        {lin2022robust}
\bibfield{author}{\bibinfo{person}{Shanchuan Lin}, \bibinfo{person}{Linjie
  Yang}, \bibinfo{person}{Imran Saleemi}, {and} \bibinfo{person}{Soumyadip
  Sengupta}.} \bibinfo{year}{2022}\natexlab{}.
\newblock \showarticletitle{Robust high-resolution video matting with temporal
  guidance}. In \bibinfo{booktitle}{\emph{Proceedings of the IEEE/CVF Winter
  Conference on Applications of Computer Vision}}. \bibinfo{pages}{238--247}.
\newblock


\bibitem[Liu et~al\mbox{.}(2020)]%
        {liu2020neural}
\bibfield{author}{\bibinfo{person}{Lingjie Liu}, \bibinfo{person}{Jiatao Gu},
  \bibinfo{person}{Kyaw Zaw~Lin}, \bibinfo{person}{Tat-Seng Chua}, {and}
  \bibinfo{person}{Christian Theobalt}.} \bibinfo{year}{2020}\natexlab{}.
\newblock \showarticletitle{Neural sparse voxel fields}.
\newblock \bibinfo{journal}{\emph{Advances in Neural Information Processing
  Systems}}  \bibinfo{volume}{33} (\bibinfo{year}{2020}),
  \bibinfo{pages}{15651--15663}.
\newblock


\bibitem[Lorensen and Cline(1998)]%
        {lorensen1998marching}
\bibfield{author}{\bibinfo{person}{William~E Lorensen} {and}
  \bibinfo{person}{Harvey~E Cline}.} \bibinfo{year}{1998}\natexlab{}.
\newblock \showarticletitle{Marching cubes: A high resolution 3D surface
  construction algorithm}.
\newblock In \bibinfo{booktitle}{\emph{Seminal graphics: pioneering efforts
  that shaped the field}}. \bibinfo{pages}{347--353}.
\newblock


\bibitem[Lu et~al\mbox{.}(2020)]%
        {lu2020improved}
\bibfield{author}{\bibinfo{person}{Conny Lu}, \bibinfo{person}{Praneeth
  Chakravarthula}, \bibinfo{person}{Yujie Tao}, \bibinfo{person}{Steven Chen},
  {and} \bibinfo{person}{Henry Fuchs}.} \bibinfo{year}{2020}\natexlab{}.
\newblock \showarticletitle{Improved vergence and accommodation via purkinje
  image tracking with multiple cameras for ar glasses}. In
  \bibinfo{booktitle}{\emph{2020 IEEE International Symposium on Mixed and
  Augmented Reality (ISMAR)}}. IEEE, \bibinfo{pages}{320--331}.
\newblock


\bibitem[Ma et~al\mbox{.}(2022)]%
        {ma2022neural}
\bibfield{author}{\bibinfo{person}{Li Ma}, \bibinfo{person}{Xiaoyu Li},
  \bibinfo{person}{Jing Liao}, \bibinfo{person}{Xuan Wang}, \bibinfo{person}{Qi
  Zhang}, \bibinfo{person}{Jue Wang}, {and} \bibinfo{person}{Pedro~V Sander}.}
  \bibinfo{year}{2022}\natexlab{}.
\newblock \showarticletitle{Neural parameterization for dynamic human head
  editing}.
\newblock \bibinfo{journal}{\emph{ACM Transactions on Graphics (TOG)}}
  \bibinfo{volume}{41}, \bibinfo{number}{6} (\bibinfo{year}{2022}),
  \bibinfo{pages}{1--15}.
\newblock


\bibitem[Ma et~al\mbox{.}(2024)]%
        {ma20243d}
\bibfield{author}{\bibinfo{person}{Shengjie Ma}, \bibinfo{person}{Yanlin Weng},
  \bibinfo{person}{Tianjia Shao}, {and} \bibinfo{person}{Kun Zhou}.}
  \bibinfo{year}{2024}\natexlab{}.
\newblock \showarticletitle{3D Gaussian Blendshapes for Head Avatar Animation}.
  In \bibinfo{booktitle}{\emph{ACM SIGGRAPH 2024 Conference Papers}}.
  \bibinfo{pages}{1--10}.
\newblock


\bibitem[Mildenhall et~al\mbox{.}(2021)]%
        {mildenhall2021nerf}
\bibfield{author}{\bibinfo{person}{Ben Mildenhall}, \bibinfo{person}{Pratul~P
  Srinivasan}, \bibinfo{person}{Matthew Tancik}, \bibinfo{person}{Jonathan~T
  Barron}, \bibinfo{person}{Ravi Ramamoorthi}, {and} \bibinfo{person}{Ren Ng}.}
  \bibinfo{year}{2021}\natexlab{}.
\newblock \showarticletitle{Nerf: Representing scenes as neural radiance fields
  for view synthesis}.
\newblock \bibinfo{journal}{\emph{Commun. ACM}} \bibinfo{volume}{65},
  \bibinfo{number}{1} (\bibinfo{year}{2021}), \bibinfo{pages}{99--106}.
\newblock


\bibitem[M{\"u}ller et~al\mbox{.}(2022)]%
        {muller2022instant}
\bibfield{author}{\bibinfo{person}{Thomas M{\"u}ller}, \bibinfo{person}{Alex
  Evans}, \bibinfo{person}{Christoph Schied}, {and} \bibinfo{person}{Alexander
  Keller}.} \bibinfo{year}{2022}\natexlab{}.
\newblock \showarticletitle{Instant neural graphics primitives with a
  multiresolution hash encoding}.
\newblock \bibinfo{journal}{\emph{ACM Transactions on Graphics (ToG)}}
  \bibinfo{volume}{41}, \bibinfo{number}{4} (\bibinfo{year}{2022}),
  \bibinfo{pages}{1--15}.
\newblock


\bibitem[Neog et~al\mbox{.}(2016)]%
        {neog2016interactive}
\bibfield{author}{\bibinfo{person}{Debanga~R Neog}, \bibinfo{person}{Jo{\~a}o~L
  Cardoso}, \bibinfo{person}{Anurag Ranjan}, {and} \bibinfo{person}{Dinesh~K
  Pai}.} \bibinfo{year}{2016}\natexlab{}.
\newblock \showarticletitle{Interactive gaze driven animation of the eye
  region}. In \bibinfo{booktitle}{\emph{Proceedings of the 21st International
  Conference on Web3D Technology}}. \bibinfo{pages}{51--59}.
\newblock


\bibitem[Nishino and Nayar(2004)]%
        {nishino2004world}
\bibfield{author}{\bibinfo{person}{Ko Nishino} {and} \bibinfo{person}{Shree~K
  Nayar}.} \bibinfo{year}{2004}\natexlab{}.
\newblock \showarticletitle{The world in an eye [eye image interpretation]}. In
  \bibinfo{booktitle}{\emph{Proceedings of the 2004 IEEE Computer Society
  Conference on Computer Vision and Pattern Recognition, 2004. CVPR 2004.}},
  Vol.~\bibinfo{volume}{1}. IEEE, \bibinfo{pages}{I--I}.
\newblock


\bibitem[Park et~al\mbox{.}(2019)]%
        {park2019deepsdf}
\bibfield{author}{\bibinfo{person}{Jeong~Joon Park}, \bibinfo{person}{Peter
  Florence}, \bibinfo{person}{Julian Straub}, \bibinfo{person}{Richard
  Newcombe}, {and} \bibinfo{person}{Steven Lovegrove}.}
  \bibinfo{year}{2019}\natexlab{}.
\newblock \showarticletitle{Deepsdf: Learning continuous signed distance
  functions for shape representation}. In \bibinfo{booktitle}{\emph{Proceedings
  of the IEEE/CVF conference on computer vision and pattern recognition}}.
  \bibinfo{pages}{165--174}.
\newblock


\bibitem[Park et~al\mbox{.}(2021a)]%
        {park2021nerfies}
\bibfield{author}{\bibinfo{person}{Keunhong Park}, \bibinfo{person}{Utkarsh
  Sinha}, \bibinfo{person}{Jonathan~T Barron}, \bibinfo{person}{Sofien
  Bouaziz}, \bibinfo{person}{Dan~B Goldman}, \bibinfo{person}{Steven~M Seitz},
  {and} \bibinfo{person}{Ricardo Martin-Brualla}.}
  \bibinfo{year}{2021}\natexlab{a}.
\newblock \showarticletitle{Nerfies: Deformable neural radiance fields}. In
  \bibinfo{booktitle}{\emph{Proceedings of the IEEE/CVF International
  Conference on Computer Vision}}. \bibinfo{pages}{5865--5874}.
\newblock


\bibitem[Park et~al\mbox{.}(2021b)]%
        {park2021hypernerf}
\bibfield{author}{\bibinfo{person}{Keunhong Park}, \bibinfo{person}{Utkarsh
  Sinha}, \bibinfo{person}{Peter Hedman}, \bibinfo{person}{Jonathan~T. Barron},
  \bibinfo{person}{Sofien Bouaziz}, \bibinfo{person}{Dan~B Goldman},
  \bibinfo{person}{Ricardo Martin-Brualla}, {and} \bibinfo{person}{Steven~M.
  Seitz}.} \bibinfo{year}{2021}\natexlab{b}.
\newblock \showarticletitle{HyperNeRF: A Higher-Dimensional Representation for
  Topologically Varying Neural Radiance Fields}.
\newblock \bibinfo{journal}{\emph{ACM Trans. Graph.}} \bibinfo{volume}{40},
  \bibinfo{number}{6}, Article \bibinfo{articleno}{238} (\bibinfo{date}{dec}
  \bibinfo{year}{2021}).
\newblock


\bibitem[Pumarola et~al\mbox{.}(2021)]%
        {pumarola2021d}
\bibfield{author}{\bibinfo{person}{Albert Pumarola}, \bibinfo{person}{Enric
  Corona}, \bibinfo{person}{Gerard Pons-Moll}, {and} \bibinfo{person}{Francesc
  Moreno-Noguer}.} \bibinfo{year}{2021}\natexlab{}.
\newblock \showarticletitle{D-nerf: Neural radiance fields for dynamic scenes}.
  In \bibinfo{booktitle}{\emph{Proceedings of the IEEE/CVF Conference on
  Computer Vision and Pattern Recognition}}. \bibinfo{pages}{10318--10327}.
\newblock


\bibitem[Qian et~al\mbox{.}(2024)]%
        {qian2024gaussianavatars}
\bibfield{author}{\bibinfo{person}{Shenhan Qian}, \bibinfo{person}{Tobias
  Kirschstein}, \bibinfo{person}{Liam Schoneveld}, \bibinfo{person}{Davide
  Davoli}, \bibinfo{person}{Simon Giebenhain}, {and} \bibinfo{person}{Matthias
  Nie{\ss}ner}.} \bibinfo{year}{2024}\natexlab{}.
\newblock \showarticletitle{Gaussianavatars: Photorealistic head avatars with
  rigged 3d gaussians}. In \bibinfo{booktitle}{\emph{Proceedings of the
  IEEE/CVF Conference on Computer Vision and Pattern Recognition}}.
  \bibinfo{pages}{20299--20309}.
\newblock


\bibitem[Rosu and Behnke(2023)]%
        {rosu2023permutosdf}
\bibfield{author}{\bibinfo{person}{Radu~Alexandru Rosu} {and}
  \bibinfo{person}{Sven Behnke}.} \bibinfo{year}{2023}\natexlab{}.
\newblock \showarticletitle{Permutosdf: Fast multi-view reconstruction with
  implicit surfaces using permutohedral lattices}. In
  \bibinfo{booktitle}{\emph{Proceedings of the IEEE/CVF Conference on Computer
  Vision and Pattern Recognition}}. \bibinfo{pages}{8466--8475}.
\newblock


\bibitem[Schwartz et~al\mbox{.}(2020)]%
        {schwartz2020eyes}
\bibfield{author}{\bibinfo{person}{Gabriel Schwartz}, \bibinfo{person}{Shih-En
  Wei}, \bibinfo{person}{Te-Li Wang}, \bibinfo{person}{Stephen Lombardi},
  \bibinfo{person}{Tomas Simon}, \bibinfo{person}{Jason Saragih}, {and}
  \bibinfo{person}{Yaser Sheikh}.} \bibinfo{year}{2020}\natexlab{}.
\newblock \showarticletitle{The eyes have it: An integrated eye and face model
  for photorealistic facial animation}.
\newblock \bibinfo{journal}{\emph{ACM Transactions on Graphics (TOG)}}
  \bibinfo{volume}{39}, \bibinfo{number}{4} (\bibinfo{year}{2020}),
  \bibinfo{pages}{91--1}.
\newblock


\bibitem[Seitz and Dyer(1999)]%
        {seitz1999photorealistic}
\bibfield{author}{\bibinfo{person}{Steven~M Seitz} {and}
  \bibinfo{person}{Charles~R Dyer}.} \bibinfo{year}{1999}\natexlab{}.
\newblock \showarticletitle{Photorealistic scene reconstruction by voxel
  coloring}.
\newblock \bibinfo{journal}{\emph{International Journal of Computer Vision}}
  \bibinfo{volume}{35} (\bibinfo{year}{1999}), \bibinfo{pages}{151--173}.
\newblock


\bibitem[Shao et~al\mbox{.}(2023)]%
        {shao2023tensor4d}
\bibfield{author}{\bibinfo{person}{Ruizhi Shao}, \bibinfo{person}{Zerong
  Zheng}, \bibinfo{person}{Hanzhang Tu}, \bibinfo{person}{Boning Liu},
  \bibinfo{person}{Hongwen Zhang}, {and} \bibinfo{person}{Yebin Liu}.}
  \bibinfo{year}{2023}\natexlab{}.
\newblock \showarticletitle{Tensor4d: Efficient neural 4d decomposition for
  high-fidelity dynamic reconstruction and rendering}. In
  \bibinfo{booktitle}{\emph{Proceedings of the IEEE/CVF Conference on Computer
  Vision and Pattern Recognition}}. \bibinfo{pages}{16632--16642}.
\newblock


\bibitem[Sun et~al\mbox{.}(2022)]%
        {sun2022direct}
\bibfield{author}{\bibinfo{person}{Cheng Sun}, \bibinfo{person}{Min Sun}, {and}
  \bibinfo{person}{Hwann-Tzong Chen}.} \bibinfo{year}{2022}\natexlab{}.
\newblock \showarticletitle{Direct voxel grid optimization: Super-fast
  convergence for radiance fields reconstruction}. In
  \bibinfo{booktitle}{\emph{Proceedings of the IEEE/CVF Conference on Computer
  Vision and Pattern Recognition}}. \bibinfo{pages}{5459--5469}.
\newblock


\bibitem[Sun et~al\mbox{.}(2015)]%
        {sun2015real}
\bibfield{author}{\bibinfo{person}{Li Sun}, \bibinfo{person}{Zicheng Liu},
  {and} \bibinfo{person}{Ming-Ting Sun}.} \bibinfo{year}{2015}\natexlab{}.
\newblock \showarticletitle{Real time gaze estimation with a consumer depth
  camera}.
\newblock \bibinfo{journal}{\emph{Information Sciences—Informatics and
  Computer Science, Intelligent Systems, Applications: An International
  Journal}} \bibinfo{volume}{320}, \bibinfo{number}{C} (\bibinfo{year}{2015}),
  \bibinfo{pages}{346--360}.
\newblock


\bibitem[Takikawa et~al\mbox{.}(2021)]%
        {takikawa2021neural}
\bibfield{author}{\bibinfo{person}{Towaki Takikawa}, \bibinfo{person}{Joey
  Litalien}, \bibinfo{person}{Kangxue Yin}, \bibinfo{person}{Karsten Kreis},
  \bibinfo{person}{Charles Loop}, \bibinfo{person}{Derek Nowrouzezahrai},
  \bibinfo{person}{Alec Jacobson}, \bibinfo{person}{Morgan McGuire}, {and}
  \bibinfo{person}{Sanja Fidler}.} \bibinfo{year}{2021}\natexlab{}.
\newblock \showarticletitle{Neural geometric level of detail: Real-time
  rendering with implicit 3d shapes}. In \bibinfo{booktitle}{\emph{Proceedings
  of the IEEE/CVF Conference on Computer Vision and Pattern Recognition}}.
  \bibinfo{pages}{11358--11367}.
\newblock


\bibitem[Thies et~al\mbox{.}(2016)]%
        {thies2016face2face}
\bibfield{author}{\bibinfo{person}{Justus Thies}, \bibinfo{person}{Michael
  Zollhofer}, \bibinfo{person}{Marc Stamminger}, \bibinfo{person}{Christian
  Theobalt}, {and} \bibinfo{person}{Matthias Nie{\ss}ner}.}
  \bibinfo{year}{2016}\natexlab{}.
\newblock \showarticletitle{Face2face: Real-time face capture and reenactment
  of rgb videos}. In \bibinfo{booktitle}{\emph{Proceedings of the IEEE
  conference on computer vision and pattern recognition}}.
  \bibinfo{pages}{2387--2395}.
\newblock


\bibitem[Tretschk et~al\mbox{.}(2021)]%
        {tretschk2021non}
\bibfield{author}{\bibinfo{person}{Edgar Tretschk}, \bibinfo{person}{Ayush
  Tewari}, \bibinfo{person}{Vladislav Golyanik}, \bibinfo{person}{Michael
  Zollh{\"o}fer}, \bibinfo{person}{Christoph Lassner}, {and}
  \bibinfo{person}{Christian Theobalt}.} \bibinfo{year}{2021}\natexlab{}.
\newblock \showarticletitle{Non-rigid neural radiance fields: Reconstruction
  and novel view synthesis of a dynamic scene from monocular video}. In
  \bibinfo{booktitle}{\emph{Proceedings of the IEEE/CVF International
  Conference on Computer Vision}}. \bibinfo{pages}{12959--12970}.
\newblock


\bibitem[Verbin et~al\mbox{.}(2022)]%
        {verbin2022ref}
\bibfield{author}{\bibinfo{person}{Dor Verbin}, \bibinfo{person}{Peter Hedman},
  \bibinfo{person}{Ben Mildenhall}, \bibinfo{person}{Todd Zickler},
  \bibinfo{person}{Jonathan~T Barron}, {and} \bibinfo{person}{Pratul~P
  Srinivasan}.} \bibinfo{year}{2022}\natexlab{}.
\newblock \showarticletitle{Ref-nerf: Structured view-dependent appearance for
  neural radiance fields}. In \bibinfo{booktitle}{\emph{2022 IEEE/CVF
  Conference on Computer Vision and Pattern Recognition (CVPR)}}. IEEE,
  \bibinfo{pages}{5481--5490}.
\newblock


\bibitem[Von~Helmholtz(1925)]%
        {von1925helmholtz}
\bibfield{author}{\bibinfo{person}{Hermann Von~Helmholtz}.}
  \bibinfo{year}{1925}\natexlab{}.
\newblock \bibinfo{booktitle}{\emph{Helmholtz's treatise on physiological
  optics}}. Vol.~\bibinfo{volume}{3}.
\newblock \bibinfo{publisher}{Optical Society of America}.
\newblock


\bibitem[Wang and Ji(2017)]%
        {wang2017real}
\bibfield{author}{\bibinfo{person}{Kang Wang} {and} \bibinfo{person}{Qiang
  Ji}.} \bibinfo{year}{2017}\natexlab{}.
\newblock \showarticletitle{Real time eye gaze tracking with 3d deformable
  eye-face model}. In \bibinfo{booktitle}{\emph{Proceedings of the IEEE
  International Conference on Computer Vision}}. \bibinfo{pages}{1003--1011}.
\newblock


\bibitem[Wang et~al\mbox{.}(2021)]%
        {wang2021neus}
\bibfield{author}{\bibinfo{person}{Peng Wang}, \bibinfo{person}{Lingjie Liu},
  \bibinfo{person}{Yuan Liu}, \bibinfo{person}{Christian Theobalt},
  \bibinfo{person}{Taku Komura}, {and} \bibinfo{person}{Wenping Wang}.}
  \bibinfo{year}{2021}\natexlab{}.
\newblock \showarticletitle{NeuS: Learning Neural Implicit Surfaces by Volume
  Rendering for Multi-view Reconstruction}.
\newblock \bibinfo{journal}{\emph{Advances in Neural Information Processing
  Systems}}  \bibinfo{volume}{34} (\bibinfo{year}{2021}),
  \bibinfo{pages}{27171--27183}.
\newblock


\bibitem[Wang et~al\mbox{.}(2023)]%
        {wang2023neus2}
\bibfield{author}{\bibinfo{person}{Yiming Wang}, \bibinfo{person}{Qin Han},
  \bibinfo{person}{Marc Habermann}, \bibinfo{person}{Kostas Daniilidis},
  \bibinfo{person}{Christian Theobalt}, {and} \bibinfo{person}{Lingjie Liu}.}
  \bibinfo{year}{2023}\natexlab{}.
\newblock \showarticletitle{Neus2: Fast learning of neural implicit surfaces
  for multi-view reconstruction}. In \bibinfo{booktitle}{\emph{Proceedings of
  the IEEE/CVF International Conference on Computer Vision}}.
  \bibinfo{pages}{3295--3306}.
\newblock


\bibitem[Wen et~al\mbox{.}(2020)]%
        {wen2020accurate}
\bibfield{author}{\bibinfo{person}{Quan Wen}, \bibinfo{person}{Derek Bradley},
  \bibinfo{person}{Thabo Beeler}, \bibinfo{person}{Seonwook Park},
  \bibinfo{person}{Otmar Hilliges}, \bibinfo{person}{Junhai Yong}, {and}
  \bibinfo{person}{Feng Xu}.} \bibinfo{year}{2020}\natexlab{}.
\newblock \showarticletitle{Accurate Real-time 3D Gaze Tracking Using a
  Lightweight Eyeball Calibration}. In \bibinfo{booktitle}{\emph{Computer
  Graphics Forum}}, Vol.~\bibinfo{volume}{39}. Wiley Online Library,
  \bibinfo{pages}{475--485}.
\newblock


\bibitem[Wen et~al\mbox{.}(2017)]%
        {wen2017real}
\bibfield{author}{\bibinfo{person}{Quan Wen}, \bibinfo{person}{Feng Xu},
  \bibinfo{person}{Ming Lu}, {and} \bibinfo{person}{Jun-Hai Yong}.}
  \bibinfo{year}{2017}\natexlab{}.
\newblock \showarticletitle{Real-time 3D eyelids tracking from semantic edges}.
\newblock \bibinfo{journal}{\emph{ACM Transactions on Graphics (TOG)}}
  \bibinfo{volume}{36}, \bibinfo{number}{6} (\bibinfo{year}{2017}),
  \bibinfo{pages}{1--11}.
\newblock


\bibitem[Wood et~al\mbox{.}(2016a)]%
        {wood20163d}
\bibfield{author}{\bibinfo{person}{Erroll Wood}, \bibinfo{person}{Tadas
  Baltru{\v{s}}aitis}, \bibinfo{person}{Louis-Philippe Morency},
  \bibinfo{person}{Peter Robinson}, {and} \bibinfo{person}{Andreas Bulling}.}
  \bibinfo{year}{2016}\natexlab{a}.
\newblock \showarticletitle{A 3d morphable eye region model for gaze
  estimation}. In \bibinfo{booktitle}{\emph{Computer Vision--ECCV 2016: 14th
  European Conference, Amsterdam, The Netherlands, October 11--14, 2016,
  Proceedings, Part I 14}}. Springer, \bibinfo{pages}{297--313}.
\newblock


\bibitem[Wood et~al\mbox{.}(2016b)]%
        {wood2016learning}
\bibfield{author}{\bibinfo{person}{Erroll Wood}, \bibinfo{person}{Tadas
  Baltru{\v{s}}aitis}, \bibinfo{person}{Louis-Philippe Morency},
  \bibinfo{person}{Peter Robinson}, {and} \bibinfo{person}{Andreas Bulling}.}
  \bibinfo{year}{2016}\natexlab{b}.
\newblock \showarticletitle{Learning an appearance-based gaze estimator from
  one million synthesised images}. In \bibinfo{booktitle}{\emph{Proceedings of
  the Ninth Biennial ACM Symposium on Eye Tracking Research \& Applications}}.
  \bibinfo{pages}{131--138}.
\newblock


\bibitem[Xiang et~al\mbox{.}(2024)]%
        {xiang2024flashavatar}
\bibfield{author}{\bibinfo{person}{Jun Xiang}, \bibinfo{person}{Xuan Gao},
  \bibinfo{person}{Yudong Guo}, {and} \bibinfo{person}{Juyong Zhang}.}
  \bibinfo{year}{2024}\natexlab{}.
\newblock \showarticletitle{FlashAvatar: High-fidelity Head Avatar with
  Efficient Gaussian Embedding}. In \bibinfo{booktitle}{\emph{Proceedings of
  the IEEE/CVF Conference on Computer Vision and Pattern Recognition}}.
  \bibinfo{pages}{1802--1812}.
\newblock


\bibitem[Xu et~al\mbox{.}(2024)]%
        {xu2024gaussian}
\bibfield{author}{\bibinfo{person}{Yuelang Xu}, \bibinfo{person}{Benwang Chen},
  \bibinfo{person}{Zhe Li}, \bibinfo{person}{Hongwen Zhang},
  \bibinfo{person}{Lizhen Wang}, \bibinfo{person}{Zerong Zheng}, {and}
  \bibinfo{person}{Yebin Liu}.} \bibinfo{year}{2024}\natexlab{}.
\newblock \showarticletitle{Gaussian head avatar: Ultra high-fidelity head
  avatar via dynamic gaussians}. In \bibinfo{booktitle}{\emph{Proceedings of
  the IEEE/CVF Conference on Computer Vision and Pattern Recognition}}.
  \bibinfo{pages}{1931--1941}.
\newblock


\bibitem[Xu et~al\mbox{.}(2023)]%
        {xu2023avatarmav}
\bibfield{author}{\bibinfo{person}{Yuelang Xu}, \bibinfo{person}{Lizhen Wang},
  \bibinfo{person}{Xiaochen Zhao}, \bibinfo{person}{Hongwen Zhang}, {and}
  \bibinfo{person}{Yebin Liu}.} \bibinfo{year}{2023}\natexlab{}.
\newblock \showarticletitle{Avatarmav: Fast 3d head avatar reconstruction using
  motion-aware neural voxels}. In \bibinfo{booktitle}{\emph{ACM SIGGRAPH 2023
  Conference Proceedings}}. \bibinfo{pages}{1--10}.
\newblock


\bibitem[Yariv et~al\mbox{.}(2021)]%
        {yariv2021volume}
\bibfield{author}{\bibinfo{person}{Lior Yariv}, \bibinfo{person}{Jiatao Gu},
  \bibinfo{person}{Yoni Kasten}, {and} \bibinfo{person}{Yaron Lipman}.}
  \bibinfo{year}{2021}\natexlab{}.
\newblock \showarticletitle{Volume rendering of neural implicit surfaces}.
\newblock \bibinfo{journal}{\emph{Advances in Neural Information Processing
  Systems}}  \bibinfo{volume}{34} (\bibinfo{year}{2021}),
  \bibinfo{pages}{4805--4815}.
\newblock


\bibitem[Yariv et~al\mbox{.}(2023)]%
        {yariv2023bakedsdf}
\bibfield{author}{\bibinfo{person}{Lior Yariv}, \bibinfo{person}{Peter Hedman},
  \bibinfo{person}{Christian Reiser}, \bibinfo{person}{Dor Verbin},
  \bibinfo{person}{Pratul~P Srinivasan}, \bibinfo{person}{Richard Szeliski},
  \bibinfo{person}{Jonathan~T Barron}, {and} \bibinfo{person}{Ben Mildenhall}.}
  \bibinfo{year}{2023}\natexlab{}.
\newblock \showarticletitle{Bakedsdf: Meshing neural sdfs for real-time view
  synthesis}. In \bibinfo{booktitle}{\emph{ACM SIGGRAPH 2023 Conference
  Proceedings}}. \bibinfo{pages}{1--9}.
\newblock


\bibitem[Yu et~al\mbox{.}(2024)]%
        {Yu2024GOF}
\bibfield{author}{\bibinfo{person}{Zehao Yu}, \bibinfo{person}{Torsten
  Sattler}, {and} \bibinfo{person}{Andreas Geiger}.}
  \bibinfo{year}{2024}\natexlab{}.
\newblock \showarticletitle{Gaussian Opacity Fields: Efficient High-quality
  Compact Surface Reconstruction in Unbounded Scenes}.
\newblock \bibinfo{journal}{\emph{arXiv:2404.10772}} (\bibinfo{year}{2024}).
\newblock


\bibitem[Zhao et~al\mbox{.}(2023)]%
        {zhao2023havatar}
\bibfield{author}{\bibinfo{person}{Xiaochen Zhao}, \bibinfo{person}{Lizhen
  Wang}, \bibinfo{person}{Jingxiang Sun}, \bibinfo{person}{Hongwen Zhang},
  \bibinfo{person}{Jinli Suo}, {and} \bibinfo{person}{Yebin Liu}.}
  \bibinfo{year}{2023}\natexlab{}.
\newblock \showarticletitle{Havatar: High-fidelity head avatar via facial model
  conditioned neural radiance field}.
\newblock \bibinfo{journal}{\emph{ACM Transactions on Graphics}}
  \bibinfo{volume}{43}, \bibinfo{number}{1} (\bibinfo{year}{2023}),
  \bibinfo{pages}{1--16}.
\newblock


\bibitem[Zheng et~al\mbox{.}(2022)]%
        {zheng2022avatar}
\bibfield{author}{\bibinfo{person}{Yufeng Zheng},
  \bibinfo{person}{Victoria~Fern{\'a}ndez Abrevaya}, \bibinfo{person}{Marcel~C
  B{\"u}hler}, \bibinfo{person}{Xu Chen}, \bibinfo{person}{Michael~J Black},
  {and} \bibinfo{person}{Otmar Hilliges}.} \bibinfo{year}{2022}\natexlab{}.
\newblock \showarticletitle{Im avatar: Implicit morphable head avatars from
  videos}. In \bibinfo{booktitle}{\emph{Proceedings of the IEEE/CVF Conference
  on Computer Vision and Pattern Recognition}}. \bibinfo{pages}{13545--13555}.
\newblock


\bibitem[Zheng et~al\mbox{.}(2023)]%
        {zheng2023pointavatar}
\bibfield{author}{\bibinfo{person}{Yufeng Zheng}, \bibinfo{person}{Wang Yifan},
  \bibinfo{person}{Gordon Wetzstein}, \bibinfo{person}{Michael~J Black}, {and}
  \bibinfo{person}{Otmar Hilliges}.} \bibinfo{year}{2023}\natexlab{}.
\newblock \showarticletitle{Pointavatar: Deformable point-based head avatars
  from videos}. In \bibinfo{booktitle}{\emph{Proceedings of the IEEE/CVF
  Conference on Computer Vision and Pattern Recognition}}.
  \bibinfo{pages}{21057--21067}.
\newblock


\bibitem[Zielonka et~al\mbox{.}(2023)]%
        {zielonka2023instant}
\bibfield{author}{\bibinfo{person}{Wojciech Zielonka}, \bibinfo{person}{Timo
  Bolkart}, {and} \bibinfo{person}{Justus Thies}.}
  \bibinfo{year}{2023}\natexlab{}.
\newblock \showarticletitle{Instant volumetric head avatars}. In
  \bibinfo{booktitle}{\emph{Proceedings of the IEEE/CVF Conference on Computer
  Vision and Pattern Recognition}}. \bibinfo{pages}{4574--4584}.
\newblock


\end{thebibliography}

\clearpage

\appendix

\section{Details of Data Preprocess}
Given an RGB input, we first get the face mask by the video matting method \cite{lin2022robust} and detect eye landmarks, including eyelid and iris landmarks, by the commercial service of SenseTime. Then, we generate the eyelid and iris mask by filling the polygon enclosed by detected landmarks. The iris mask is used for eyeball calibration, while the eyelid mask is used to generate the final mask for eyelid reconstruction.

\begin{figure}[htp]
	\centering
	\includegraphics[width=1.0\linewidth]{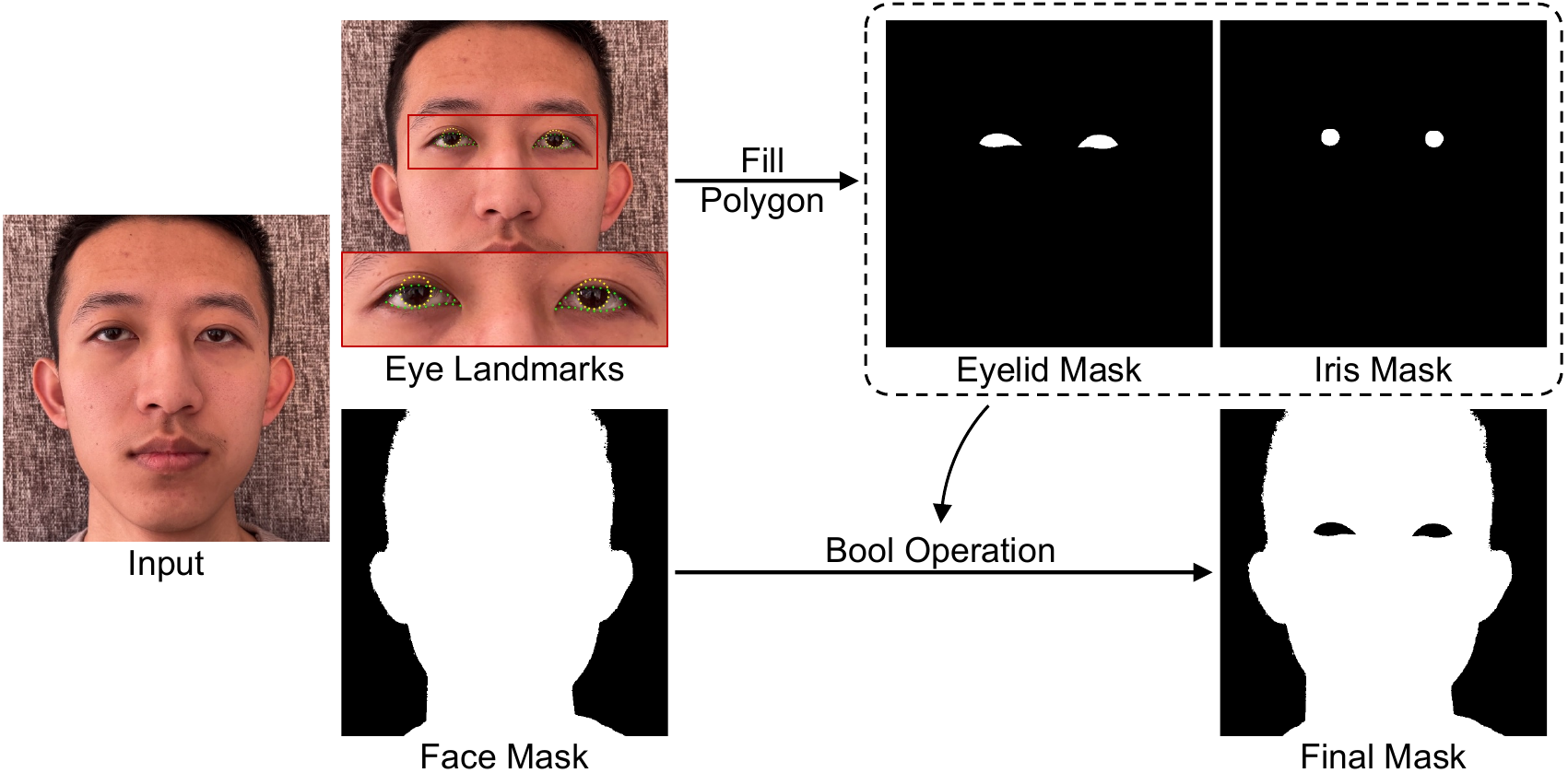}
	\caption{Mask generation. The final mask used for eyelid reconstruction excludes the eyeball region by the bool operation.}
	\label{fig:preprocess}
\end{figure}

\section{Details of Eyeball Calibration}
In order to acquire good initialization for the iris alignment, we adopt a coarse-to-fine strategy by first optimizing the eyeball position and scale by aligning the iris center and radius while keeping eyeball rotations fixed:
\begin{equation}
	\label{eq:init}
	E(\textbf{P}^{e},s;\textbf{P}^{c}_i,\mathcal{T},\mathcal{M})=\sum_{i}^{n}{\|\textbf{o}_{i}-\hat{\textbf{o}}_i\|_1}+\sum_{i}^{n}{\|r_{i}-\hat{r}_i\|_1}
\end{equation}
where $\textbf{o}\in \mathbb{R}^2$ is the 2D center of the ground-truth iris mask. $r\in \mathbb{R}$ is the 2D bounding box size (i.e., $\max(width,height)$) of the ground-truth iris mask. The symbol with hat represents the results of the rendered iris mask. Based on the coarse results, we take eyeball rotations into the optimization and change the objective to aligning the iris mask. Note that Eq. \ref{eq:init} also participate in this stage for regularization. So, the full objective is formulated as
\begin{equation}
	\begin{aligned}
		E(\textbf{P}^{e},\textbf{R}^{e}_{i},s;\textbf{P}^{c}_{i},\mathcal{T},\mathcal{M}) &= \sum_{i}^{n}{\|\mathcal{M}_i-\hat{\mathcal{M}}_i\|_2^2}\\
		&+	\lambda_1\sum_{i}^{n}{\|\textbf{o}_{i}-\hat{\textbf{o}}_i\|_1}+\lambda_2\sum_{i}^{n}{\|r_{i}-\hat{r}_i\|_1}
	\end{aligned}
\end{equation}
\begin{equation}
	\hat{\mathcal{M}}_i=\Pi { (\mathcal{T},s,\textbf{P}^{e},\textbf{P}^{c}_{i},\textbf{R}^{e}_{i})}
\end{equation}
where we use $\lambda_1=0.1$ and $\lambda_2=0.01$ in our implementation. The whole calibration costs about ten minutes on one NVIDIA RTX 3090.

\section{Disentanglement for Both Eyes}
To extend the disentanglement strategy from \{eye, others\} to \{left eye, right eye, others\}, we further split the eye code $\boldsymbol{\varphi}^e_i$ into $\boldsymbol{\varphi}^{le}_i$ and $\boldsymbol{\varphi}^{re}_i$ for the left and right eye, respectively. The strategy is depicted in Tab. \ref{tab:det_strategy}. $B^{le}$ and $B^{re}$ denote the bounding boxes of the left and right eye, respectively. $B^o$ is the other region. 

\begin{table}[htp]
	\caption{Disentanglement strategy.}
	\label{tab:det_strategy}
	\begin{tabular}{ccc|ccc}
		\hline
		\multicolumn{3}{c|}{change} & \multicolumn{3}{c}{keep same} \\ \hline
		$\boldsymbol{\varphi}^{le}_i$       & $\boldsymbol{\varphi}^{re}_i$       & $\boldsymbol{\varphi}^{o}_i$       & $B^{le}$        & $B^{re}$        & $B^{o}$       \\ \hline
		&         &  $\checkmark$       &  $\checkmark$        &  $\checkmark$        &     \\
		$\checkmark$     &         &         &          & $\checkmark$        & $\checkmark$       \\
		&  $\checkmark$       &         &  $\checkmark$        &          &  $\checkmark$       \\
		\hline   
	\end{tabular}
\end{table}

\begin{figure*}[!t]
	\centering
	\includegraphics[width=0.9\linewidth]{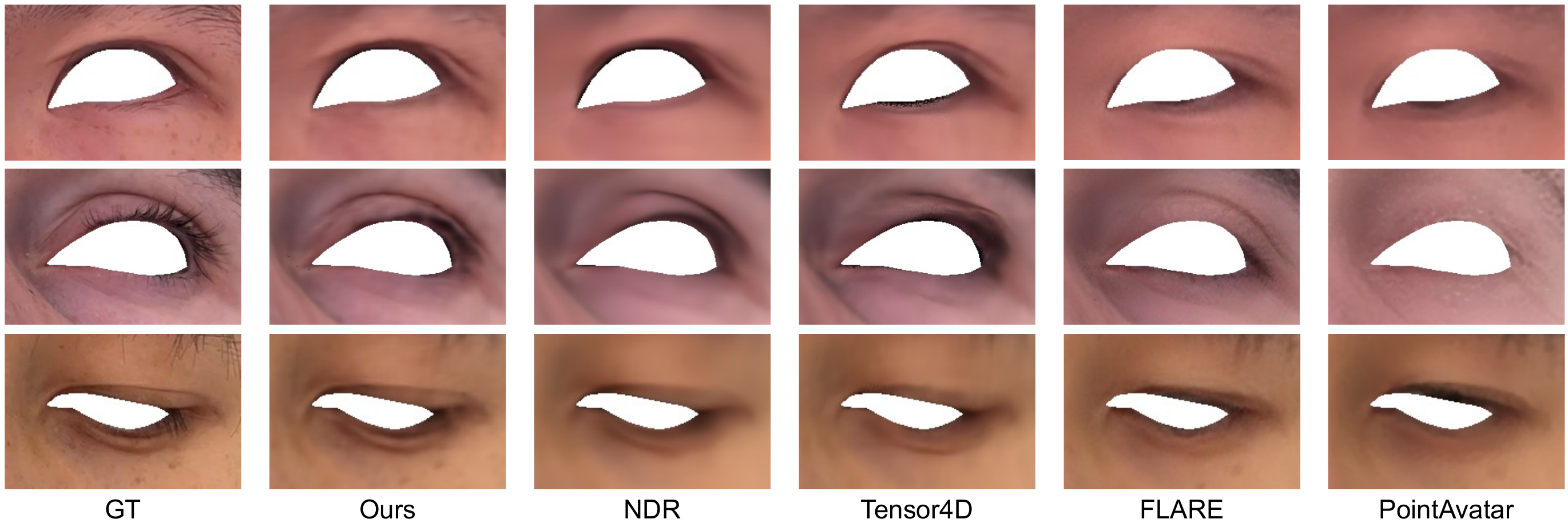}
	\caption{Rendered results on real data. Benefiting from the geometry improvements, our method outperforms other methods with the same-level capture setups.}
	\label{fig:render_results}
\end{figure*}

For example, when changing the left eye code $\boldsymbol{\varphi}^{le}_i$, the hyper-space coordinates of points in the $B^{re}$ and $B^{o}$ are encouraged to keep the same as before. Specifically, the disentangle loss is extended as:
\begin{equation}
	\begin{aligned}
		\mathcal{L}_{det}&=\frac{1}{N_1}\sum_{\textbf{X}^o\in B_1}\|H(\textbf{X}^o,[\boldsymbol{\varphi}^{le}_i,\boldsymbol{\varphi}^{re}_i,\boldsymbol{\varepsilon}])-H(\textbf{X}^o,[\boldsymbol{\varphi}^{le}_i,\boldsymbol{\varphi}^{re}_i,\boldsymbol{\varphi}^o_i])\|_1\\
		&+\frac{1}{N_2}\sum_{\textbf{X}^o\in B_2}\|H(\textbf{X}^o,[\boldsymbol{\varepsilon},\boldsymbol{\varphi}^{re}_i,\boldsymbol{\varphi}^o_i])-H(\textbf{X}^o,[\boldsymbol{\varphi}^{le}_i,\boldsymbol{\varphi}^{re}_i,\boldsymbol{\varphi}^o_i])\|_1\\
		&+\frac{1}{N_3}\sum_{\textbf{X}^o\in B_3}\|H(\textbf{X}^o,[\boldsymbol{\varphi}^{le}_i,\boldsymbol{\varepsilon},\boldsymbol{\varphi}^o_i])-H(\textbf{X}^o,[\boldsymbol{\varphi}^{le}_i,\boldsymbol{\varphi}^{re}_i,\boldsymbol{\varphi}^o_i])\|_1
	\end{aligned}
\end{equation}
where $H(\textbf{X}^o,\boldsymbol{\varphi}_i)=[F_d(\textbf{X}^o,\boldsymbol{\varphi}_i),F_w(\textbf{X}^o,\boldsymbol{\varphi}_i)]$ and $\boldsymbol{\varepsilon}$ is a random latent code sampled by standard normal distribution. $B_1=B^{le} \cup B^{re}$, $B_2=B^{re} \cup B^{o}$, and $B_3=B^{le} \cup B^{o}$ are the sets of points in the corresponding regions, while $N_1$, $N_2$, and $N_3$ are the corresponding number of points. 

\begin{figure}[htp]
	\centering
	\includegraphics[width=1.0\linewidth]{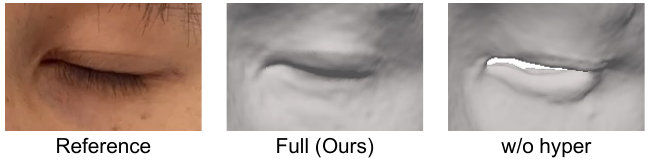}
	\caption{Ablation for ablation. Without hyper-space, the algorithm cannot model the closed eye.}
	\label{fig:ablation_hyper}
\end{figure}

\section{Modeling of Closing Eyes}
Closing eyes is an important movement for realistic eyelid animation but cannot be directly controlled by eyeball rotations. In order to model and control it, we additionally introduce two latent codes of 32 dimensions (named closing code) to encode the closing conditions, representing closed and non-closed eyes, respectively.
According to the eyelid masks, we can automatically obtain the closing condition of each frame by setting a threshold of the pixel distance between upper and lower eyelids, and we feed the corresponding closing code into the deformation network and topology network during training.
When inserting closing eyes into an animation sequence, we linearly interpolate the closing code of closed eyes with that of non-closed eyes.

\section{Motivation of Hyper-space}
The hyper-space is designed to handle the topology changes during eyelids moving, especially when participants closing their eyes. To demonstrate its benefits, we add another ablation study that removes hyper-space and only model the deformation by the deformation network $F_d$. Results in Fig. \ref{fig:ablation_hyper} show that without hyper-space, the algorithm fails to model the closed eyes. 
Apart from topology changes, the ability of geometry modeling provided by the hyper-space also benefits the geometry reconstruction of non-closed eyelids. We compare geometry metrics of two experiment settings on our synthetic dataset. Results in Tab. \ref{tab:hyperspace} show the advantages of our full method.

\begin{table}[htp]
	\footnotesize
	\caption{Ablation study of hyper-space.}
	\label{tab:hyperspace}
	\begin{tabular}{l|c|c|c|c|c|c|c|c}
		\hline
		\multirow{2}{*}{Methods} & \multicolumn{4}{c|}{Depth Error $\downarrow$} & \multicolumn{4}{c}{Chamfer Distance $\downarrow$} \\ \cline{2-9} 
		& ID-1         & ID-2        & ID-3        & ID-4  & ID-1         & ID-2        & ID-3        & ID-4      \\
		\hline
		w/o hyper          &      1.13    &      0.72    &     1.20     &    1.04    &    0.112     &       0.071   &    0.153      &     0.119   \\
		Full (Ours)           &      \textbf{0.79}     &     \textbf{0.48}     &    \textbf{0.60}     &     \textbf{0.49} &      \textbf{0.044}     &     \textbf{0.038}     &    \textbf{0.062}     &     \textbf{0.034}   \\
		\hline
	\end{tabular}
\end{table}

\section{Rendering Quality}
Apart from geometry, our method can also obtain colors from the appearance network $F_c$. We compare the rendering quality of different methods both qualitatively and quantitatively. As shown in Fig. \ref{fig:render_results}, our method show better rendering quality in the eye region. Specifically, PointAvatar \cite{zheng2023pointavatar} and FLARE \cite{bharadwaj2023flare} suffer from the misalignment of wrinkles, while NDR \cite{cai2022neural} and Tensor4D \cite{shao2023tensor4d} miss some details. The mean PSNR of all participants in Tab. \ref{tab:render_cmp} also demonstrate our advantages.

\begin{table}[htp]
	\caption{Mean PSNR of different methods in the eye region.}
	\label{tab:render_cmp}
	\begin{tabular}{ccccc}
		\hline
		Ours  & NDR   & Tensor4D & FLARE & PointAvatar  \\ \hline 
		30.29 & 28.44 & 28.19  & 21.97 & 20.86  \\ 
		\hline
	\end{tabular}
\end{table}

\section{Differences from Similar Methods}
Although the linear combination of our adaptive anchor grids is similar with some techniques like NeRFBlendShape \cite{gao2022reconstructing} and NeRSemble \cite{kirschstein2023nersemble} in formulation, there are some in-depth differences among these methods, and our method has some advantages. 
First, the impact of linear combination in NeRFBlendShape \cite{gao2022reconstructing} is different from NeRSemble \cite{kirschstein2023nersemble} and ours. 
As they do not use canonical plus deformation architecture, the linear combination in NeRFBlendShape \cite{gao2022reconstructing} aims for capabilities of representing various expressions rather than expression-dependent encodings of features.  
Second, the combination coefficients of NeRSemble \cite{kirschstein2023nersemble} are learnable parameters, while ours are semantic parameters (i.e., eyeball rotations) that are fixed once acquired, which is more suitable for animation control. As for reconstruction, we present an ablation that changes our eyeball rotations to learnable parameters just as NeRSemble \cite{kirschstein2023nersemble}. The quantitative results shown in Tab. \ref{tab:learnable_gaze} show that using learnable parameters has no obvious advantages on geometry reconstruction.

\begin{table}[htp]
	\footnotesize
	\caption{Ablation study of learnable parameters.}
	\label{tab:learnable_gaze}
	\begin{tabular}{l|c|c|c|c|c|c|c|c}
		\hline
		\multirow{2}{*}{Methods} & \multicolumn{4}{c|}{Depth Error $\downarrow$} & \multicolumn{4}{c}{Chamfer Distance $\downarrow$} \\ \cline{2-9} 
		& ID-1         & ID-2        & ID-3        & ID-4  & ID-1         & ID-2        & ID-3        & ID-4      \\
		\hline
		learnable gaze           &      1.12    &     1.20    &     0.72      &    0.80    &      0.071    &       0.198    &     \textbf{0.053}       &    0.050    \\
		Full (Ours)           &      \textbf{0.79}     &     \textbf{0.48}     &    \textbf{0.60}     &     \textbf{0.49} &      \textbf{0.044}     &     \textbf{0.038}     &    0.062     &     \textbf{0.034}   \\
		\hline
	\end{tabular}
\end{table}

\section{Quantitative Comparisons of Other Facial Regions}
Our method models the other facial regions together with eyelids in the same SDF field. The difference is that the other facial regions are only modeled by the naive combination of NDR \cite{cai2022neural} and NeuDA \cite{cai2023neuda}, while the eye region benefits from the proposed techniques (contact loss and gaze-dependent adaptive anchor grid). In order to fully show the advantages of our method, we supplement the geometry metrics of the other facial regions on synthetic data. As shown in Tab. \ref{tab:geo_cmp_other}, we compare our method with the two most competitive methods (NDR \cite{cai2022neural} and Tensor4D \cite{shao2023tensor4d}). Our method also achieve the best geometry reconstruction.

\begin{table}[htp]
	\footnotesize
	\caption{Quantitative comparisons of other facial regions on synthetic data.}
	\label{tab:geo_cmp_other}
	\begin{tabular}{l|c|c|c|c|c|c|c|c}
		\hline
		\multirow{2}{*}{Methods} & \multicolumn{4}{c|}{Depth Error $\downarrow$} & \multicolumn{4}{c}{Chamfer Distance $\downarrow$} \\ \cline{2-9} 
		& ID-1         & ID-2        & ID-3        & ID-4  & ID-1         & ID-2        & ID-3        & ID-4      \\
		\hline
		NDR           &     0.50     &   0.64     &     0.51     &   0.77    &     0.050     &     0.092    &   0.042    &    0.109  \\
		Tensor4D    &      1.02     &    1.41     &       1.15    &   1.06   &     0.149     &    0.418     &    0.235     &   0.183    \\
		Ours          &      \textbf{0.35}      &     \textbf{0.40}     &     \textbf{0.39}   &    \textbf{0.44}   &     \textbf{0.040}    &    \textbf{0.029}     &     \textbf{0.025}   &    \textbf{0.078}    \\
		\hline
	\end{tabular}
\end{table}

\section{Interpolation Behaviors}
\begin{figure}[htp]
	\centering
	\includegraphics[width=1.0\linewidth]{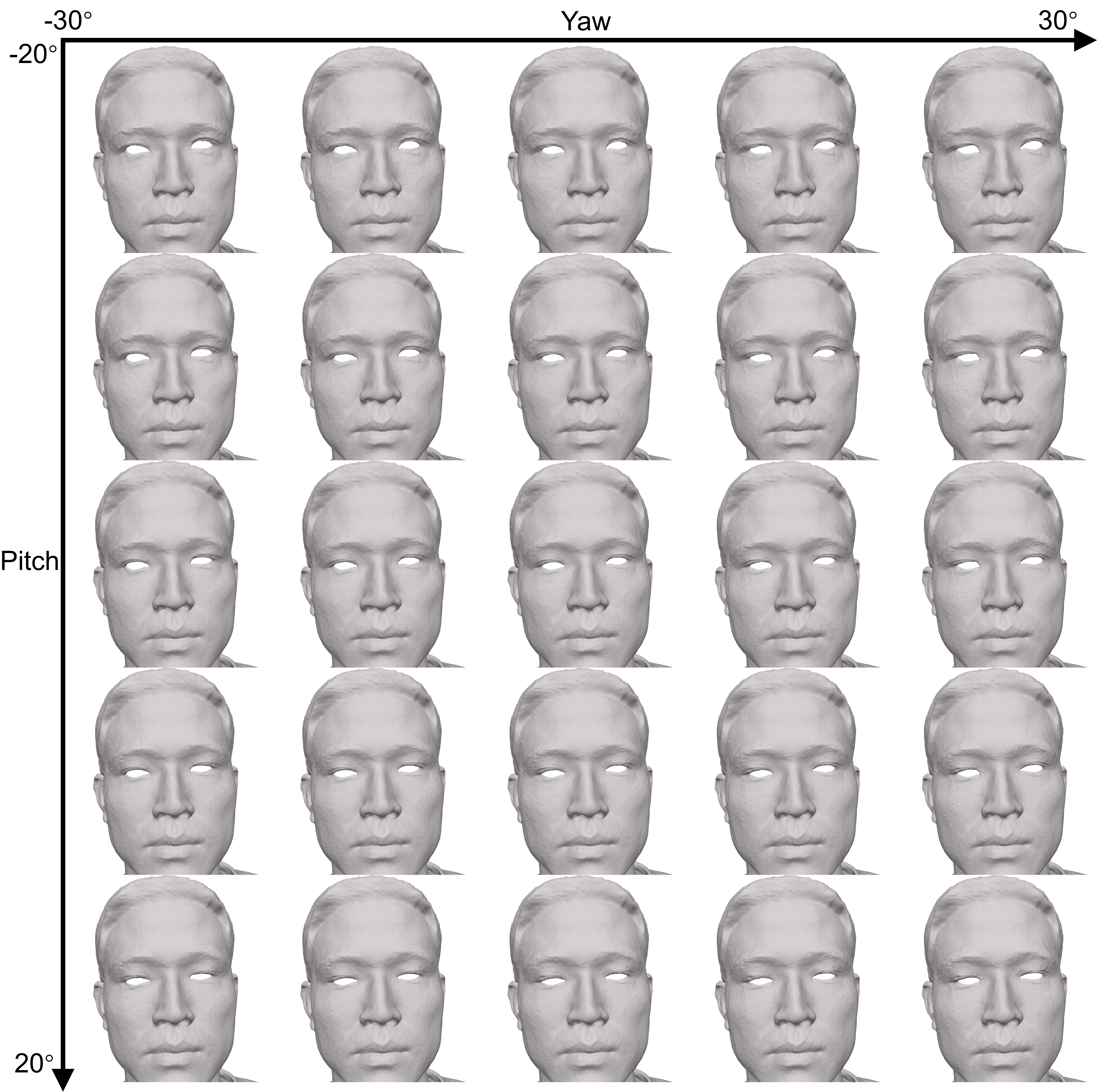}
	\caption{Visualization of interpolation behaviors. Our reconstructed eyelids can be animated by eyeball rotations.}
	\label{fig:interpolation_grid}
\end{figure}

Our reconstructed eyelids can be animated by semantic parameters (i.e., eyeball rotations). The valid range of rotation values vary from person to person because different people can rotate their eyeballs at different angles. In general, most case works well with a pitch angle from $-20^{\circ}$ to $20^{\circ}$ and a yaw angle from $-30^{\circ}$ to $30^{\circ}$ (an example is shown in Fig. \ref{fig:interpolation_grid}). A larger angle may result in artifacts if it is beyond the range of eyeball rotations in the training data.

\section{Tolerance of Calibration Errors}
\begin{figure}[!t]
	\centering
	\includegraphics[width=1.0\linewidth]{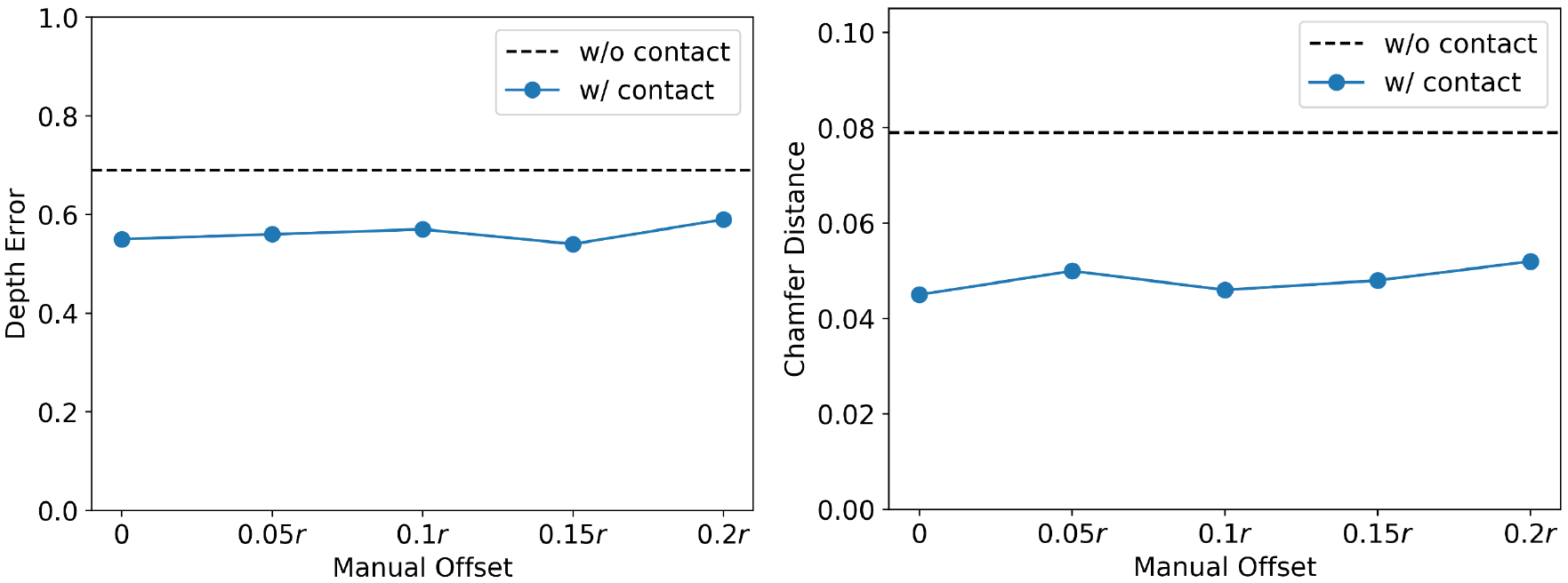}
	\caption{Trend of reconstruction performance changing with calibration errors. The x-axis is the length (represented by the ratio of eyeball radius) of manual offsets. Using inaccurate parameters to construct contact loss is still beneficial for eyelid reconstruction and outperforms the w/o contact ablation.}
	\label{fig:tolerance}
\end{figure}

As our eyelid reconstruction relies on eyeball parameters from eyeball calibration, it is valuable to explore its robustness to calibration errors. 
However, as we do not have ground-truth eyeball parameters, it is infeasible to get the calibration errors directly. Instead, we make calibration errors manually by adding man-made offsets to the eyeball parameters of our full method. 
Considering the key to eyeball calibration is to determine the eyeball positions, this experiment mainly focuses on the robustness to position errors.
Specifically, the added manual offset is a vector that contains a random direction with a ratio of eyeball radius. For each ID in our synthetic data, we randomly sample the direction and uniformly sample the ratio from 0.05 to 0.2. The mean depth error and Chamfer distance of setups with different manual offsets are used to reflect the trend of reconstruction performance changing with position errors. 
As shown in Fig. \ref{fig:tolerance}, with the manual offset increases, the reconstruction performance of setups with non-zero manual offsets is still comparable with that of the original setup (0 manual offset). In addition, even with a large offset of 0.2 times the radius length, using the inaccurate parameters to construct contact loss is still beneficial for eyelid reconstruction, outperforming the w/o contact ablation. To some extent, these phenomena demonstrate that our method is robust to calibration errors.

\section{More Results on Synthetic Data}
\begin{figure*}[!t]
	\centering
	\includegraphics[width=1.0\linewidth]{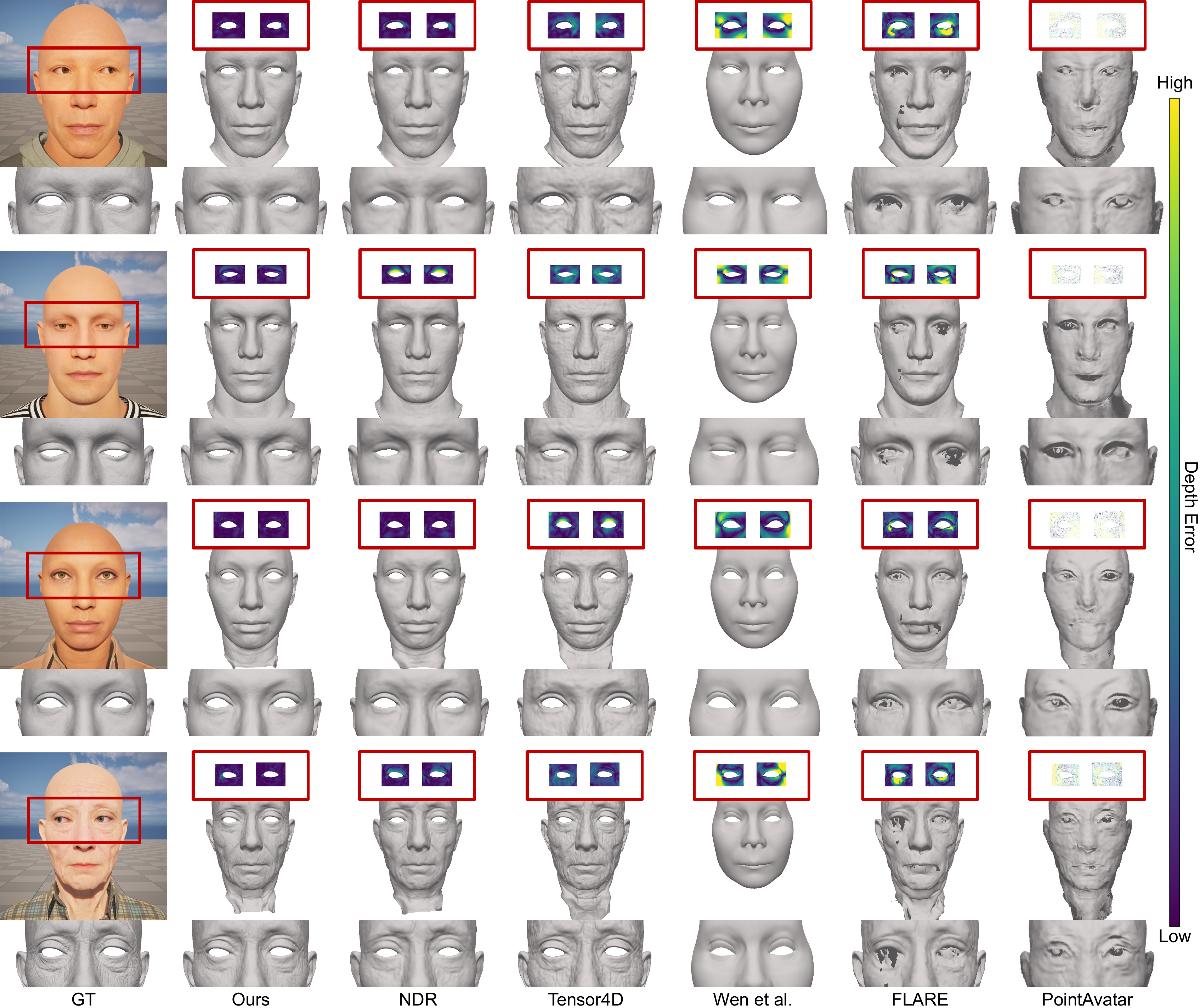}
	\caption{Reconstructed results on synthetic data with corresponding depth error map of the eye region.}
	\label{fig:compare}
\end{figure*}

We show qualitative comparisons on synthetic data in Fig. \ref{fig:compare}. Similar to the results on real data, PointAvatar \cite{zheng2023pointavatar} can only reconstruct coarse contours of facial details, and the results of \citet{wen2017real} are seriously restricted to the model capacity. FLARE \cite{bharadwaj2023flare} can reconstruct some personal details, but the results have severe self-intersection. The results of Tensor4D \cite{shao2023tensor4d} contain some bumpy artifacts, while the results of NDR \cite{cai2022neural} miss some subtle details. Our method outputs more plausible geometry and reconstructs more subtle details in the eye region.
The depth error maps in the red rectangle also demonstrate the advantages of our method. Note that PointAvatar \cite{zheng2023pointavatar} is a point-based method and thus only contains sparse values on the error maps.

\end{document}